\documentclass[fleqn,10pt]{wlscirep}
\usepackage[utf8]{inputenc}
\usepackage{graphicx}
\usepackage{diagbox}
\usepackage{booktabs}
\usepackage{multirow}
\usepackage{array}
\usepackage{amsmath, amssymb}
\usepackage{bm}

\DeclareMathOperator{\F}{F}
\DeclareMathOperator{\T}{T}
\newcommand{\norm}[1]{\lVert#1\rVert}

\title{Hyperspectral Imaging}

\author[1,$\ddag$,*]{Danfeng Hong}
\author[1,$\ddag$]{Chenyu Li}
\author[2]{Naoto Yokoya}
\author[3]{Bing Zhang}
\author[4]{Xiuping Jia}
\author[5]{Antonio Plaza}
\author[6]{Paolo Gamba}
\author[7]{Jon Atli Benediktsson}
\author[8]{Jocelyn Chanussot}
\affil[1]{Southeast University, Nanjing 210096, China}
\affil[2]{Graduate School of Frontier Sciences, the University of Tokyo, Chiba 277-8561, Japan}
\affil[3]{Aerospace Information Research Institute, Chinese Academy of Sciences, Beijing 100094, China}
\affil[4]{School of Engineering Technology, University of New South Wales, Canberra, ACT 2612, Australia}
\affil[5]{Department of Technology of Computers and Communications, Escuela Polit\'ecnica, University of Extremadura, 10003 C\'aceres, Spain}
\affil[6]{Department of Electrical, Computer and Biomedical Engineering, University of Pavia, Pavia 27100, Italy}
\affil[7]{Faculty of Electrical and Computer Engineering, University of Iceland, Reykjavik 102, Iceland}
\affil[8]{Univ. Grenoble Alpes, INRIA, CNRS, Grenoble INP, LJK, Grenoble 38000, France}
\affil[$\ddag$]{Contributed equally}
\affil[*]{Corresponding author: Danfeng Hong (hongdanfeng1989@gmail.com)}

\begin{abstract}
Hyperspectral imaging (HSI) is an advanced sensing modality that simultaneously captures spatial and spectral information, enabling non-invasive, label-free analysis of material, chemical, and biological properties. This Primer presents a comprehensive overview of HSI, from the underlying physical principles and sensor architectures to key steps in data acquisition, calibration, and correction. We summarize common data structures and highlight classical and modern analysis methods, including dimensionality reduction, classification, spectral unmixing, and AI-driven techniques such as deep learning. Representative applications across Earth observation, precision agriculture, biomedicine, industrial inspection, cultural heritage, and security are also discussed, emphasizing HSI’s ability to uncover sub-visual features for advanced monitoring, diagnostics, and decision-making. Persistent challenges, such as hardware trade-offs, acquisition variability, and the complexity of high-dimensional data, are examined alongside emerging solutions, including computational imaging, physics-informed modeling, cross-modal fusion, and self-supervised learning. Best practices for dataset sharing, reproducibility, and metadata documentation are further highlighted to support transparency and reuse. Looking ahead, we explore future directions toward scalable, real-time, and embedded HSI systems, driven by sensor miniaturization, self-supervised learning, and foundation models. As HSI evolves into a general-purpose, cross-disciplinary platform, it holds promise for transformative applications in science, technology, and society. 
\end{abstract}
\begin{document}

\flushbottom
\maketitle

\thispagestyle{empty}

\section*{Introduction}
Hyperspectral imaging (HSI) is an advanced optical sensing technique that judiciously assembles spectroscopy and digital photography into a single system. This integration enables simultaneous acquisition of spatial and spectral data, capturing spatially resolved images across numerous contiguous spectral bands. As a result, each pixel in the captured scene possesses a unique spectral signature, often referred to as a ``fingerprint''.  These pixel-level spectra are structured into three-dimensional datasets, known as hyperspectral data cubes, which combine two spatial dimensions with one spectral dimension, effectively bridging conventional imaging and spectroscopy. The hyperspectral cube contains rich spectral information obtained by measuring reflectance or radiance across finely resolved spectral intervals. Such detailed spectral content enables precise identification and thorough characterization of materials, biological tissues, and environmental surfaces.

\begin{figure*}[!t]
    \centering
    \includegraphics[width=1\linewidth]{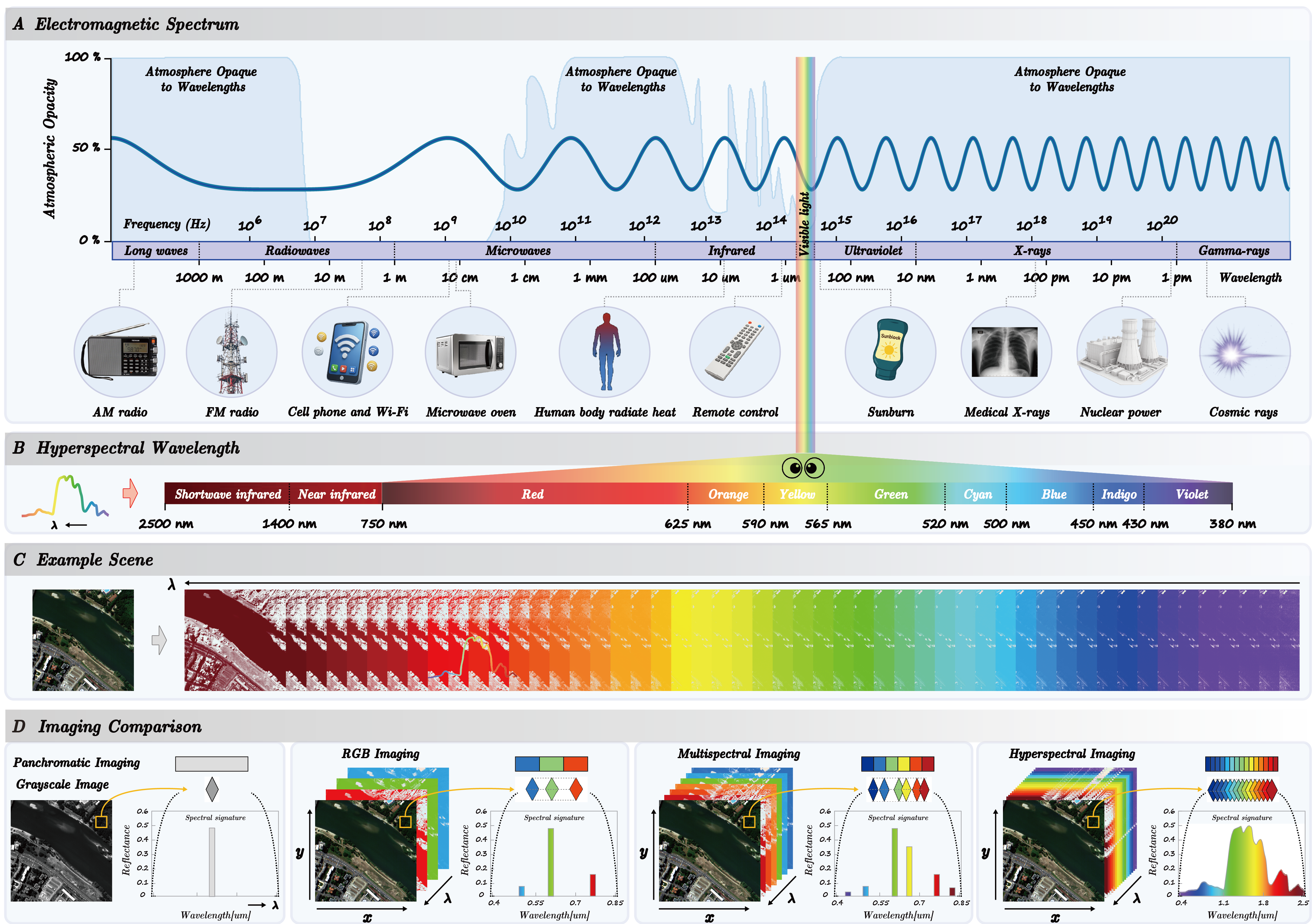}
    \caption{Overview of hyperspectral imaging. (A) Graphical illustration of the electromagnetic spectrum. (B) Expanded view of the typical wavelength regions captured in HSI: visible light (400-750 nm), near-infrared (NIR, 750-1400 nm), and shortwave infrared (SWIR, 1400-2500 nm). (C) Example hyperspectral scene illustrating integrated spatial and spectral information. (D) Qualitative spectral comparison of imaging modalities: panchromatic (single broadband), RGB (three broad bands), multispectral ($<$ 20 discrete bands), and hyperspectral ($>$ 100 contiguous narrow bands).}
    \label{fig:background}
\end{figure*}

HSI focuses on the optical window of the electromagnetic spectrum (see Fig. \ref{fig:background}A), typically covering wavelengths from 380 to 2500 nm\cite{rybicki2024radiative}. This window usually encompasses the visible light (400-700 nm), near-infrared (NIR), and shortwave infrared (SWIR) regions, as shown in Fig. \ref{fig:background}B. Notably, the inclusion of the 380-400 nm region further enhances sensitivity to pigment absorption and subtle surface reflectance variations, particularly in the violet and near-ultraviolet range. An example scene illustrating typical hyperspectral data is shown in Fig. \ref{fig:background}C, highlighting that HSI routinely captures over hundreds of spectral channels at high spectral resolution (commonly 5-10 nm). 

This extensive spectral coverage significantly differentiates HSI from panchromatic imaging, conventional color photography, and multispectral imaging. Fig. \ref{fig:background}D visualizes the spectral characteristics of different imaging modalities and makes a horizontal comparison qualitatively. Panchromatic imaging systems record a single broad spectral band encompassing the visible spectrum, resulting in high spatial resolution but minimal spectral detail. Standard RGB cameras typically record only three broad spectral bands, while multispectral systems generally capture fewer than 20 discrete bands. In contrast, the rich spectral detail provided by HSI enables researchers to identify subtle yet diagnostically significant features, including molecular absorption bands and pigment-related transitions. These features, often invisible or indistinct in conventional imaging, allow precise discrimination between targets that may appear indistinguishable in visible bands yet differ substantially in their underlying chemical composition or physiological state.
 
The high-dimensional nature of hyperspectral data stems from fundamental physical interactions between incident radiation and target materials, including electronic transitions, molecular vibrations, scattering, and fluorescence. These interactions, which vary according to the imaging modality and spectral domain, underpin the unique capability of HSI to provide non-destructive, label-free analyses. Consequently, HSI is uniquely positioned to probe not only surface characteristics but also subsurface properties, offering valuable insights across diverse scientific and practical domains.

The field of HSI has its roots in remote sensing, particularly in airborne and spaceborne Earth observation \cite{goetz1985imaging}. Early systems such as NASA's Airborne Visible/Infrared Imaging Spectrometer (AVIRIS) \cite{vane1993airborne}, developed in the 1980s, were primarily designed for mapping land cover, assessing vegetation health, and identifying mineral compositions from aircraft platforms. Over the following decades, advances in optics, detectors, and computational imaging have driven the miniaturization and diversification of HSI systems. Today, HSI devices span a wide range of factors, from satellite payloads and drone-mounted cameras to laboratory microscopes and portable point-of-care instruments, extending their applications across disciplines such as environmental science, agriculture, medicine, security, and manufacturing. A recent trend is the emergence of hyperspectral CubeSat constellations, led by companies like Kuva Space and Pixxel, which offer higher revisit rates and global coverage at reduced cost. These compact, AI-enhanced platforms are accelerating the shift from experimental demonstrations to large-scale, operational Earth observation services.

What makes HSI particularly valuable is its inherently non-destructive and label-free nature. In contrast to invasive biochemical assays or staining-based microscopy, HSI does not require physical contact or chemical labeling of the sample. It is thus suitable for \textit{in situ} and \textit{in vivo} applications, making it highly attractive for continuous monitoring, real-time diagnostics, and automation. Moreover, because HSI captures both spatial patterns and spectral fingerprints, it can be applied to tasks that range from crop disease identification and surgical guidance to counterfeit detection and geological mapping \cite{chang2003hyperspectral}. These properties position HSI as a versatile, information-rich platform for high-throughput and high-fidelity sensing.

Despite its advantages, the intrinsically high-dimensional nature of HSI data poses substantial challenges for processing and interpretation. Classical statistical methods, such as principal component analysis (PCA) \cite{greenacre2022principal} and linear discriminant analysis (LDA) \cite{zhao2024linear}, remain widely used for dimensionality reduction and feature extraction. To address spectral mixing that arises from limited spatial resolution and complex surface heterogeneity, spectral unmixing \cite{bioucas2012hyperspectral} decomposes mixed pixels into constituent endmember spectra and their fractional abundances, enabling accurate material identification and sub-pixel characterization. Increasingly, advances in machine learning and artificial intelligence (AI) are transforming the field, offering powerful tools to model the complexity and variability inherent to hyperspectral data \cite{hong2021interpretable}. Convolutional neural networks, spectral–spatial architectures, self-supervised learning, and transformer-based models have markedly advanced capabilities in classification, anomaly detection, and target localization. More recently, foundation models \cite{hong2024multimodal} trained on diverse, large-scale hyperspectral datasets have demonstrated the potential to generalize across tasks, domains, and sensor types. Together, these AI-native approaches are driving a paradigm shift in hyperspectral data exploitation, enabling scalable, automated, and robust analytical workflows.

This Primer aims to provide an accessible yet authoritative overview of the methodology behind HSI. We emphasize the current state of the field, focusing on the physical mechanisms of HSI, sensor modalities and acquisition strategies, representative data outputs, and cutting-edge analytical approaches. Applications are highlighted across diverse domains, including Earth observation, agriculture, medicine, industry, culture, and defense. In addition, we address practical challenges in reproducibility and data sharing, including the need for metadata standards and community data repositories. We also explore how emerging computational strategies, particularly those leveraging deep learning, can overcome existing limitations and enable new capabilities.

Notably, this Primer is neither intended as a technical manual nor a step-by-step guide for specific instruments, nor as an exhaustive review of every HSI hardware configuration or domain-specific protocol. Instead, our goal is to present the underlying principles and methodological considerations that define the field, while highlighting current trends and identifying opportunities for future research. By synthesizing insights across disciplines, we aim to support and encourage broader adoption, refinement, and ultimately greater impact of HSI in science and society.

\section*{Experimentation}
This section outlines the general workflow of HSI, emphasizing key stages in the HSI pipeline. Representative data examples from different imaging platforms are then categorized to clarify data characteristics. Finally, some common software tools and recommended best practices for standardized data collection are also discussed.

\subsection*{Pipeline of HSI systems}
\begin{figure*}[!t]
    \centering
    \includegraphics[width=1\linewidth]{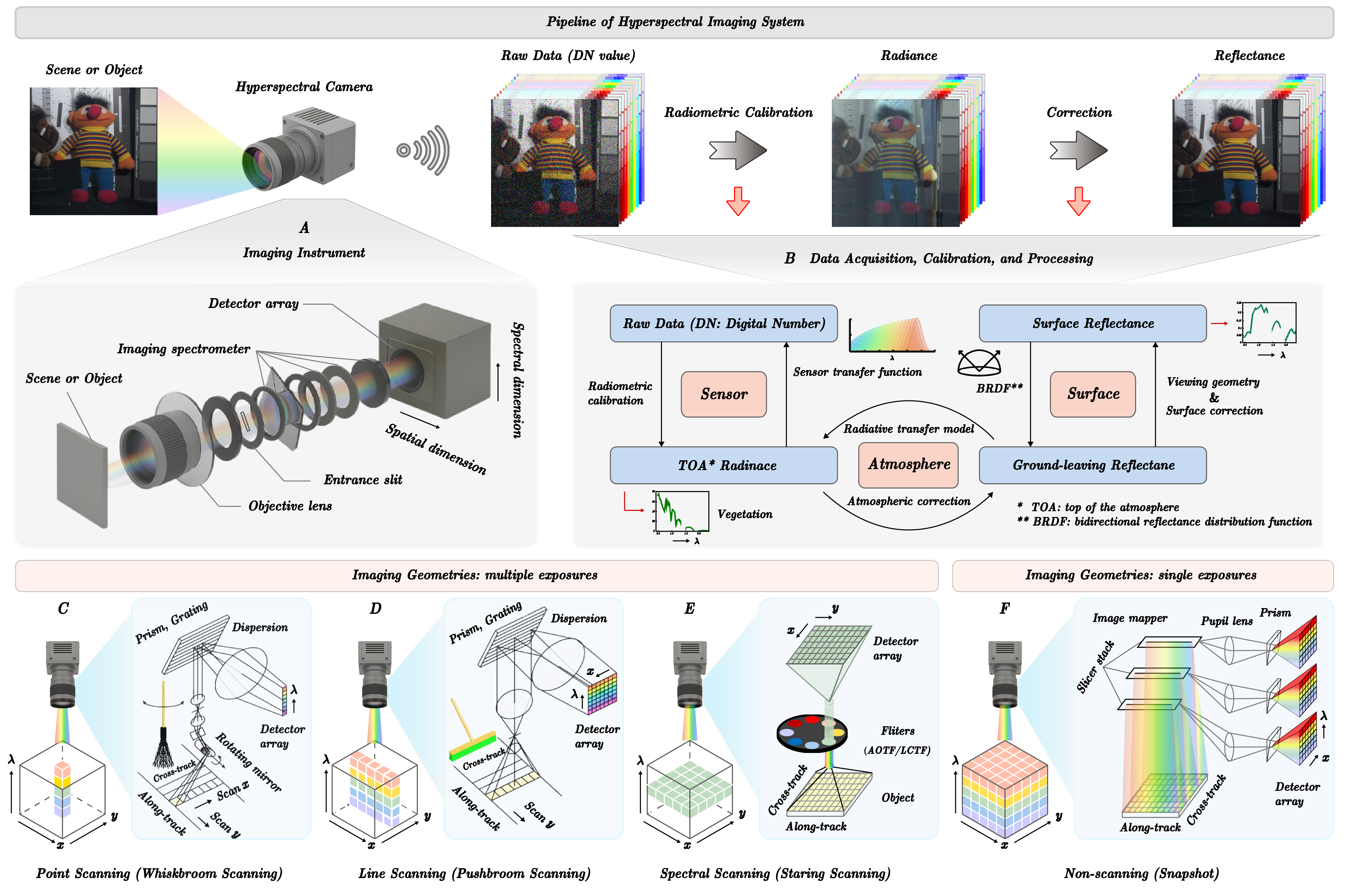}
    \caption{Pipeline of HSI systems, from the to-be-captured scene to processed reflectance data. (A) Schematic of an imaging instrument to illustrate its core components. (B) Workflow of data acquisition, calibration, and processing to clarify the conversion from raw digital number (DN) values to radiance and final reflectance data. (C)-(F) Evolution of scanning techniques: (C) whiskbroom (point) scanning, (D) pushbroom (line) scanning, (E) staring (spectral) scanning--each employing multiple exposures, and (F) snapshot imaging, acquiring all spectral bands simultaneously in a single exposure without scanning.}
    \label{fig:pipeline}
\end{figure*}

Fig. \ref{fig:pipeline} illustrates the pipeline of HSI systems, highlighting the imaging instrument and its core component, i.e., the imaging spectrometer. It further outlines a workflow encompassing data acquisition, calibration, and preprocessing, and depicts mainstream imaging geometries commonly used in HSI. In detail, a typical HSI workflow begins with the scene or object to be captured. Reflected or emitted light from this target first passes through an optical assembly, which collects and directs the incoming radiation toward the imaging spectrometer. The spectrometer then spatially separates the radiation into numerous contiguous spectral bands. These spectrally dispersed signals are subsequently captured by a detector array (or sensor), which converts optical information into measurable electrical signals. Finally, the raw data are transformed into standardized digital image data suitable for further analysis through a preprocessing phase, such as calibration and correction steps.

\subsubsection*{Optical assembly}
The optical assembly of an HSI system efficiently collects, directs, and focuses incident radiation reflected or emitted by the observed scene or object. Typically comprising lenses, mirrors, or hybrid lens-mirror combinations, this assembly establishes critical imaging parameters, such as field of view (FOV), point spread function (PSF), spatial resolution, and spectral range. High-quality optical design is crucial, as it reduces chromatic and geometric aberrations, ensuring accurate and reliable preservation of spatial details and spectral fidelity during data acquisition.

\subsubsection*{Imaging spectrometer}
The imaging spectrometer is the core component that distinguishes HSI systems from traditional optical imaging methods. Its primary function is to spectrally disperse incident radiation into numerous narrow, contiguous wavelength bands, generating detailed spectral signatures for each spatial location (or image pixel). A critical element within the imaging spectrometer responsible for this spectral dispersion is the dispersion optics module, which typically employs diffraction gratings, prisms, or electronically tunable filters. The spectrometer’s spectral resolution, defined as the ability to differentiate closely spaced wavelengths, is crucial for the accurate identification, discrimination, and characterization of materials and targets.

\noindent \textbf{Spectral dispersion techniques}. Dispersion optics employ several approaches to achieve effective spectral separation. \textit{Diffraction gratings} spatially separate wavelengths according to diffraction angles, providing high spectral resolution and excellent mechanical stability, and thus are especially suitable for airborne and satellite-based HSI systems. By contrast, \textit{prism-based systems} disperse incoming radiation through wavelength-dependent refraction. While prisms are robust and stable, their nonlinear dispersion characteristics complicate precise spectral calibration. Alternatively, \textit{electronically tunable filters}, including liquid crystal tunable filters (LCTFs)\cite{hardeberg2002multispectral} and acousto-optic tunable filters (AOTFs)\cite{chang1981acousto}, achieve spectral separation by dynamically modulating optical properties, specifically birefringence in LCTFs or acoustic interactions in AOTFs. These tunable filters offer considerable flexibility and rapid spectral selection, which are beneficial in laboratory, handheld, or stationary imaging scenarios. Therefore, the choice of a specific dispersion element directly shapes the performance, complexity, and suitable application scenarios of the imaging spectrometer and the overall HSI system.

\subsubsection*{Detector array}
The sensor or detector array captures spectrally dispersed optical radiation and converts it into measurable electronic signals. Common detector technologies in HSI include charge-coupled devices (CCD) and complementary metal-oxide semiconductor (CMOS) sensors, each with specific strengths regarding sensitivity, quantum efficiency, dynamic range, and noise characteristics. The choice of a detector technology directly influences critical system performance metrics, particularly signal-to-noise ratio (SNR), radiometric accuracy, and overall data quality. Following detection, the electrical signals are digitized and subjected to preprocessing steps, e.g., radiometric calibration, geometric correction, and noise reduction, to produce high-quality (e.g., reliable, standardized) hyperspectral data for subsequent analysis.

\subsubsection*{Imaging geometries}
In digital imaging, the concept of imaging geometry usually refers to the relationship between the object under observation and its representation in the image plane, while in HSI systems, this term specifically denotes distinct scanning modalities used for data acquisition. Four primary geometries, i.e., pushbroom, whiskbroom, snapshot, and staring, are commonly employed in practical applications. Each modality provides distinct advantages, tailored to specific observational requirements and use-case scenarios.

\noindent \textbf{Scanning techniques.} \textit{Pushbroom scanner}, also known as line-scanning, simultaneously records spectral information across an entire spatial line. As the platform, such as a satellite, aircraft, or drone, moves forward, successive spatial lines combine to form a two-dimensional image. Pushbroom systems efficiently deliver high spatial and spectral resolution, making them particularly suitable for satellite and airborne-based applications. \textit{Whiskbroom scanner}, or point-scanning, uses mechanically actuated scanning mirrors or rotating prisms to sequentially capture individual pixels. This modality offers precise radiometric accuracy and operational flexibility, but is mechanically more complex and relatively slower, limiting its widespread application compared to pushbroom systems. Whiskbroom approaches, however, remain valuable for specialized airborne and certain scientific missions. \textit{Snapshot scanners} simultaneously acquire spatial and spectral data in a single exposure without mechanical scanning. Enabled by advanced computational optical designs, including coded aperture or micro-lens arrays combined with dispersive elements, snapshot imaging is ideal for capturing rapidly changing scenes or transient events. Despite these advantages, snapshot methods typically involve trade-offs in spectral resolution and SNR. Finally, \textit{staring scanners} acquire spectral bands sequentially by employing electronically tunable filters (such as LCTFs or AOTFs). Commonly used in laboratory settings, handheld instruments, or stationary monitoring systems, staring imagers offer high spectral flexibility and accuracy. However, their extended acquisition times and need for stable imaging conditions generally limit their applicability to dynamic or rapidly changing scenes.

\subsubsection*{Data acquisition, calibration, and processing}
Following the selection of an appropriate imaging geometry and spectral distortion technique, accurate hyperspectral data acquisition necessitates systematic processing and calibration procedures to ensure data quality, comparability, and consistency before further analysis. Essential preprocessing steps generally comprise radiometric calibration, geometric correction, atmospheric correction, and wavelength calibration, each critical for achieving reliable and reproducible hyperspectral datasets.

\noindent \textbf{Calibration and correction workflows.} \textit{Radiometric calibration} converts raw sensor measurements into physically meaningful reflectance or radiance values by correcting for sensor response variability and detector sensitivity. This step usually involves imaging well-characterized reference targets, such as standardized white and dark calibration panels, to normalize and ensure consistency of the recorded signals. \textit{Geometric correction} aligns the acquired spectral data spatially, correcting for distortions due to platform motion, instrument optics, and imaging geometry. Precise geometric alignment is critical for accurately comparing and analyzing images acquired at different times or from different platforms. For airborne and satellite-based HSI platforms, \textit{atmospheric correction} is particularly important. Atmospheric constituents such as water vapor, aerosols, and atmospheric gases can absorb or scatter incoming radiation, distorting the measured spectral signatures. Atmospheric correction algorithms, often radiative transfer model-based or empirically derived, are applied to remove these influences and retrieve accurate, standardized surface reflectance values. Additionally, \textit{wavelength calibration} ensures that the spectral bands recorded by the imaging spectrometer precisely correspond to known wavelength standards. This procedure typically uses reference sources or well-defined absorption features to validate and, where required, adjust spectral alignment.

\noindent \textbf{Environmental and operational considerations.} 
Beyond these calibration procedures, environmental conditions and operational practices, including illumination conditions, weather variability, and observation angles, significantly affect the consistency, spectral integrity, and reliability of hyperspectral data acquisition. Illumination conditions, such as solar angle, atmospheric clarity, and shadowing, require careful control and thorough documentation. Ideal outdoor imaging conditions typically involve clear skies, stable illumination (e.g., midday sun), and minimal movement within the observed scene. Viewing geometry factors, such as sensor tilt, height, and target orientation, significantly affect spectral measurements, especially given the bidirectional reflectance characteristics of various surfaces.

Additional operational considerations include maintaining a high SNR, ensuring stable ambient temperature, and minimizing mechanical vibrations, all of which directly impact spectral quality. Laboratory-based acquisitions benefit substantially from stable, controlled artificial illumination and regulated environmental parameters (e.g., temperature and humidity) to improve measurement repeatability and reliability. Therefore, establishing standardized protocols that comprehensively address sensor settings, environmental conditions, viewing geometries, and preprocessing methods is essential for minimizing uncertainties and ensuring the robustness and reproducibility of hyperspectral data across studies.

\subsection*{Data examples of different imaging platforms}
\subsubsection*{Overview of imaging platforms}
HSI technologies are deployed across multiple platforms tailored to diverse scientific, industrial, and environmental needs. Generally, these platforms are categorized into four main types based on operational environments and deployment scales: spaceborne, airborne, ground-based, and portable or laboratory-based systems.

\begin{itemize}
    \item Spaceborne platforms, including hyperspectral satellites, facilitate large-scale observations of Earth's surface. Originally developed for Earth monitoring tasks such as vegetation analysis, mineral mapping, and climate studies, they are also valuable for planetary exploration by providing detailed spectral information of extraterrestrial surfaces.
    \item Airborne platforms, comprising aircraft-mounted systems and unmanned aerial vehicles (UAVs), offer high spatial resolution and operational flexibility. Such platforms are widely used in environmental monitoring, precision agriculture, infrastructure inspection, archaeological mapping, and disaster response applications.
    \item Ground-based stations consist of stationary hyperspectral instruments typically installed on towers, tripods, or fixed platforms. These are used in long-term environmental observation, calibration/validation tasks, and controlled experimental studies, where repeatable and high-precision spectral measurements are required.
    \item Portable and laboratory-based instruments, including handheld and benchtop spectrometers, provide high-resolution spectral and radiometric performance. They are especially suited for material characterization, food quality analysis, medical diagnostics (e.g., cancer detection), and cultural heritage studies, where precise, point-scale spectral accuracy is critical.
\end{itemize}
Collectively, this diverse array of platforms enables HSI to serve an extensive range of applications, from geosciences and agriculture to medicine and industry. Together, these systems provide a versatile and scalable framework, supporting comprehensive spectral analyses across multiple spatial scales, from localized \textit{in situ} measurements to expansive remote sensing applications.

\subsubsection*{Comparative summary}
HSI platforms differ markedly in spatial resolution, spectral fidelity, coverage scale, and operational flexibility. Spaceborne sensors allow for broad geographic coverage and facilitate long-term monitoring, albeit at moderate spatial resolution. Airborne platforms provide higher spatial detail with flexible acquisition, but with comparatively limited spatial extent. Ground-based systems enable continuous or targeted observations at fixed sites, making them particularly valuable for calibration, validation, and time-series studies. Portable devices and laboratory-based instruments deliver the highest spectral quality and control, which is more applicable to sample-level analysis and material characterization.

Table \ref{tab:platform_comparison} lists a comparative summary of key attributes in different imaging platforms, highlighting their respective strengths and roles in hyperspectral applications. In general, spaceborne platforms provide extensive, even global imaging coverage, though with relatively lower spatial, spectral, and temporal resolutions and limited operational flexibility. Consequently, satellite-based HSI is particularly suited to large-scale monitoring applications, but it involves substantial challenges related to data transmission, storage, compression, and processing, given the extremely high data volumes generated. Airborne platforms, such as aircraft and UAVs, are more flexible (\textit{cf.} spaceborne), yielding higher resolution attributes. This benefit enables applications, e.g., precision agriculture, forestry, and natural disaster assessment. Ground stations tend to generate higher resolution products, particularly in temporal continuity, which is commonly used for locally ecological and environmental monitoring. Laboratory-based instruments provide superior spectral fidelity and precise measurements under controlled conditions, and are widely used in biomedical research, food quality inspection, and material analysis, while handheld or portable HSI systems, empowered by advances in miniaturized optics and embedded computation, are now being deployed in field studies for plant phenotyping, contamination detection, and infrastructure diagnostics.

\begin{table*}[!t]
    \centering
    \caption{Comparative summary of different HSI platforms.}
 \resizebox{1\textwidth}{!}{
    \begin{tabular}{l|c|c|c|c}
        \toprule[1.5pt]
        \diagbox{Attributes}{Platforms} & Spaceborne (Satellite) & Airborne (Aircraft, UAV) & Ground-based (Fixed site) & Portable, Lab\\
        \midrule
         Platform illustration  &
          \begin{minipage}[c]{0.18\textwidth}
            \centering
            \includegraphics[width=0.8\linewidth]{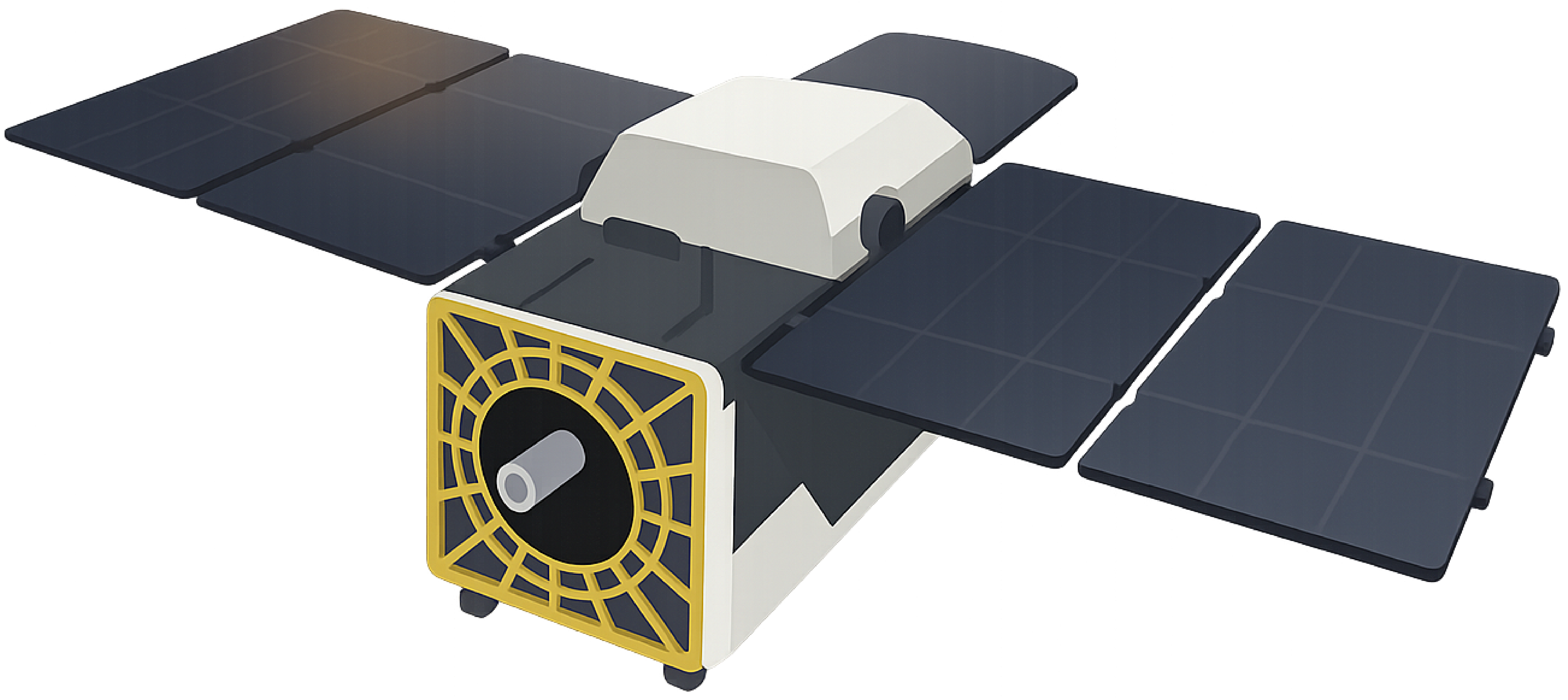}
        \end{minipage} &
        \begin{minipage}[c]{0.18\textwidth}
            \centering           \includegraphics[width=0.9\linewidth]{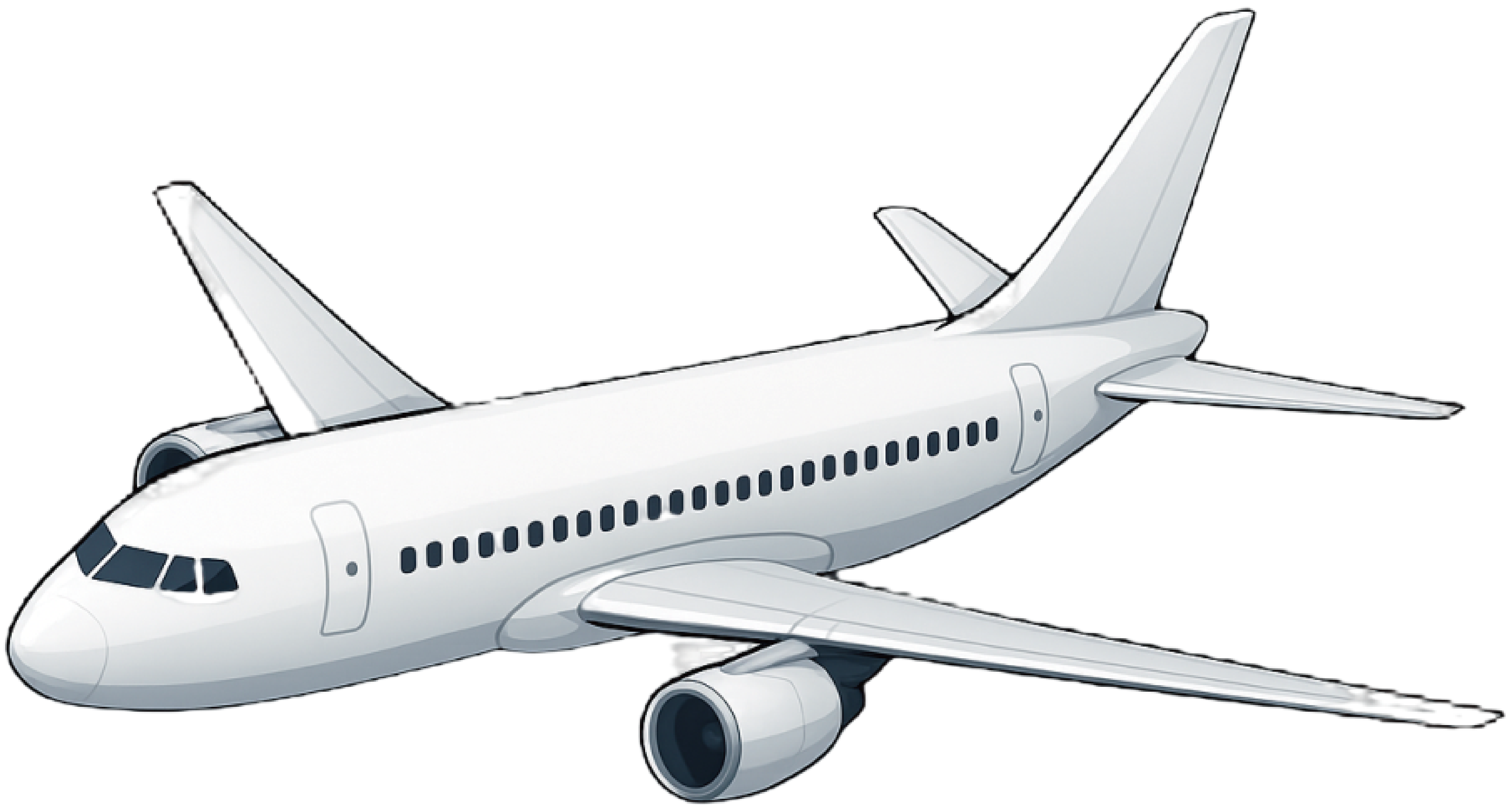}
        \end{minipage} &
        \begin{minipage}[c]{0.18\textwidth}
            \centering
            \includegraphics[width=0.65\linewidth]{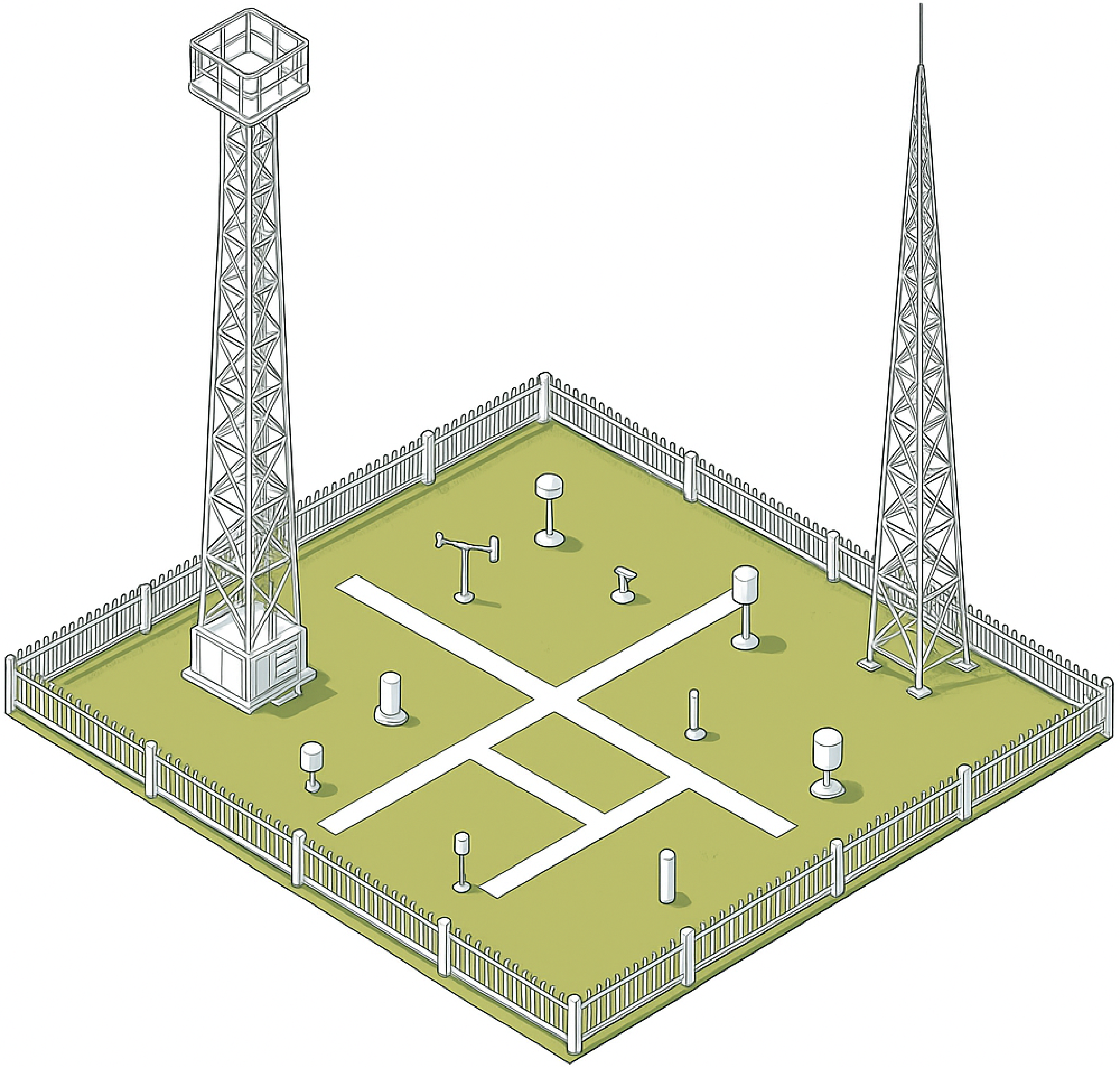}
        \end{minipage} &
        \begin{minipage}[c]{0.18\textwidth}
            \centering
            \includegraphics[width=0.75\linewidth]{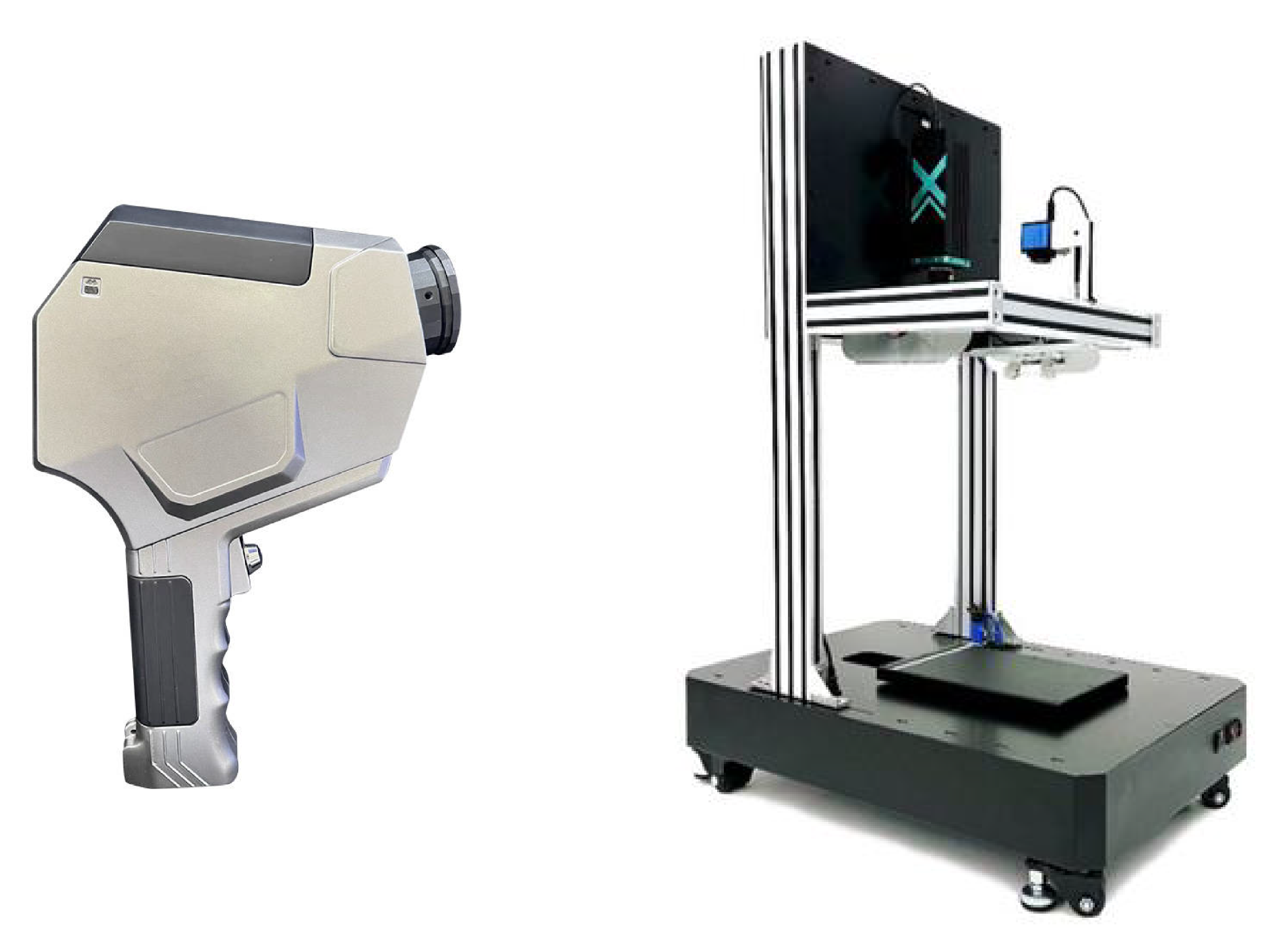}
        \end{minipage}
        \\[10pt]
        \midrule
        Spectral resolution & Medium ($\sim$5-10 nm) & High ($\sim$1-5 nm) & High ($\sim$1-3 nm) & Very high (<1 nm) \\
        Spatial resolution  & Low ($\sim$30 m) & Medium to high ($\sim$1-5 m) & High (cm to m-level) & Very high (sub-mm) \\
        Temporal resolution & Low (days to weeks) & Moderate (on-demand) & High (continuous) & Real-time controllable \\
        Coverage area & Global, continuous & Regional, flexible & Local, small area & Point-scale, lab-scale \\
        Operational flexibility & Low & Moderate to high & High & Fully controllable \\
        Imaging mode & Pushbroom, whiskbroom & Pushbroom, frame, line-scanning & Point-based, fiber-coupled & Scanning or snapshot \\
        Typical sensors & EO-1, PRISMA, EnMAP, Gaofen-5 & AVIRIS, HyMap, AISA & ASD FieldSpec, ImSpector & Resonon, Specim IQ, Headwall Nano \\
        Data volume & Very large (compressed) & Large (onboard storage) & Moderate (real-time possible) & Small (lab-efficient) \\
        Use cases & Global monitoring, agriculture, forestry & Precision agriculture, geology, disaster response & Ecological, environmental monitoring & Validation, diagnostics, lab analysis \\
        Commonality & Radiometric/spectral calibration required & Atmospheric correction needed & Standard reference spectra & Validation of other platforms \\
        \bottomrule[1.5pt]
    \end{tabular}
    }
    \label{tab:platform_comparison}
\end{table*}

To better illustrate the practical characteristics and potential applications of HSI across different platforms, representative open-access datasets are summarized in Table~\ref{tab:hsi_platform}. For instance, spaceborne datasets, collected by the Environmental Mapping and Analysis Program (EnMAP) from Germany or by PRecursore IperSpettrale della Missione Applicativa (PRISMA), the Italian hyperspectral mission, offer moderate spatial resolution suited to broad-scale monitoring, while airborne datasets, collected by AVIRIS Next Generation (NG) or NEON's Airborne Observation Platform (AOP), provide high spatial resolution for detailed regional mapping. At finer scales, ground-based libraries, collected by the United States Geological Survey (USGS) Spectroscopy Lab, specifically with Analytical Spectral Devices (ASD) FieldSpec spectroradiometers or the Ecological Spectral Information System (EcoSIS), and laboratory-based measurements, also collected using ASD FieldSpec spectroradiometers or available in SPECCHIO, a web-accessible spectral database, deliver precise spectral information critical for calibration, validation, and specialized diagnostics.

\begin{table*}[!t]
    \centering
    \caption{HSI data examples for each platform and their key characteristics.}
  \resizebox{1\textwidth}{!}{
    \begin{tabular}{l|c|c|c|c|c}
        \toprule[1.5pt]
        Platform & Dataset & Spectral bands & Spatial resolution & Typical applications & Link\\
        \hline
        \multirow{2}{*}{Spaceborne}
        & EnMAP & 242 (420-2450 nm) & 30 m & Vegetation, agriculture, forestry & \url{https://www.enmap.org/data_access/}\\
        & PRISMA & 239 (420-2450 nm) & 30 m & Geology, mineral mapping & \url{https://www.asi.it/en/earth-science/prisma/}\\
        \hline
        \multirow{2}{*}{Airborne}
        & AVIRIS-NG & 425 (380-2450 nm) & 1-5 m & Precision agriculture, ecological mapping & \url{https://avirisng.jpl.nasa.gov/dataportal/}\\
        & NEON AOP & 426 (380-2450 nm) & $\sim$1 m & Forest health, biodiversity studies & \url{https://data.neonscience.org/data-products/explore}\\
        \hline
        \multirow{2}{*}{Ground Station}
        & USGS ASD Library & $\sim$2151 (350-2500 nm) & Point-scale & Satellite calibration, vegetation modeling & \url{https://www.usgs.gov/labs/spectroscopy-lab}\\
        & EcoSIS Library & Variable (350-2500 nm) & Point-scale & Plant phenotyping, soil-water analysis & \url{https://ecosis.org/}\\
        \hline
        \multirow{2}{*}{Portable, Lab}
        & SPECCHIO Database & Variable (typ. >1000 bands) & Lab-scale & Biomedical diagnostics, material analysis & \url{https://www.specchio.ch/}\\
        & ASD FieldSpec Data & $\sim$2151 (350-2500 nm) & Lab-scale & Precise spectral characterization & \url{https://www.ipf.kit.edu/english/code_1862.php}\\
        \bottomrule[1.5pt]
    \end{tabular}
    }
    \label{tab:hsi_platform}
\end{table*}

\subsection*{Software tools and practices}
Reliable and reproducible HSI analyses depend on standardized data collection protocols and robust software tools. Multiple open-source and commercial software packages are widely employed by the HSI community, each targeting specific steps in data acquisition, calibration, preprocessing, and analysis.

Open-source tools, including \textit{HyTools} \cite{chlus2025hytools} and Python-based packages such as \textit{Spectral Python (SPy)} \cite{boggs2020spectral} and \textit{HypPy} \cite{bakker2024hyperspectral}, enable flexible, script-based workflows encompassing data preprocessing, spectral calibration, atmospheric correction, and visualization. These software packages promote transparency, reproducibility, and interoperability, thereby facilitating collaborative research. Commercial software packages, notably the environment for visualizing images (\textit{ENVI}) and \textit{ERDAS Imagine}, feature comprehensive, user-friendly interfaces and robust processing pipelines, making them particularly suitable for industrial applications and large-scale operational contexts. However, it is important to note that certain specialized software packages have been developed explicitly to address specific application domains or user requirements.

Equally critical is adherence to best practices in data collection and management, including:
\begin{itemize} 
    \item Consistent use of standardized calibration targets (e.g., white and dark panels) for accurate radiometric calibration. 
    \item Thorough documentation of metadata, such as sensor parameters, illumination conditions, and viewing geometry, ensuring accurate data interpretation and reproducibility. 
    \item Regular sensor calibration to maintain measurement consistency and reliability over extended periods. 
    \item Public dissemination of raw and processed datasets through established repositories (e.g., EcoSIS, SPECCHIO), enhancing transparency, community collaboration, and cross-validation. 
\end{itemize}

Integrating these best practices with reliable software tools is fundamental to achieving robust, reproducible hyperspectral analysis across diverse platforms and application domains.

\section*{Results}

This section emphatically introduces a general workflow of hyperspectral data analysis, as illustrated in Fig. \ref{fig:analysis}. After obtaining calibrated radiance or reflectance hyperspectral data, initial image processing steps, including restoration, enhancement, and dimensionality reduction, are typically considered. Higher-level analytical tasks, such as classification and spectral unmixing, are then performed to derive meaningful and actionable information from hyperspectral images. Furthermore, qualitative and quantitative results are also presented and interpreted.

\subsection*{Image restoration}
In the imaging process, hyperspectral data inevitably suffer from quality variations, caused by environmental conditions (e.g., illumination, temperature, humidity, and atmospheric effects) and instrumental factors (e.g., sensor noise, calibration drift, and optical aberrations). These resulting degradations, such as complex noise, striping, missing, damage, or occlusion (see Fig. \ref{fig:analysis}(A)) can significantly impact data accuracy, spectral fidelity, and subsequent analytical reliability. Therefore, image restoration is a critical step in the hyperspectral data analysis pipeline \cite{hong2021interpretable}. This process can be mathematically formulated in a general format as follows:
\begin{equation}
\label{eq1}
\begin{aligned}
       \mathbf{Y}=\mathcal{H}(\mathbf{X})+\mathbf{N},
\end{aligned}
\end{equation} 
where $\mathbf{Y}$ and $\mathbf{X}$ denote the observed degraded hyperspectral data and the unknown original clean data, respectively, while $\mathcal{H}$ stands for the degradation operator (e.g., striping, occlusion, missing, blurring, etc.), and $\mathbf{N}$ is the additive noise term, modeling various noise sources, such as sensor noise and atmospheric interference. Based on the general mathematical formulation in Eq. (\ref{eq1}), image restoration can be formulated as the following inverse problem:
\begin{equation}
\label{eq2}
\begin{aligned}     \mathbf{X}=\mathop{\arg}\mathop{\min}\limits_{\mathbf{X}} \underbrace{\mathcal{L}(\mathcal{H}(\mathbf{X}),\mathbf{Y})}_{\text{fidelity}} +\lambda \cdot \underbrace{\mathcal{R}(\mathbf{X})}_{\text{regularizer}},
\end{aligned}
\end{equation} 
where $\mathcal{L}$ denotes the data fidelity term, typically chosen as the mean square error, e.g., $\norm{\mathcal{H}(\mathbf{X})-\mathbf{Y}}^{2}_{\F}$. The regularizer $\mathcal{R}$, weighted by the regularization parameter $\lambda$, enforces specific prior assumptions such as smoothness, sparsity, or deep prior. Additionally, the degraded operator $\mathcal{H}$ varies according to the specific restoration task. Examples include: 1) denoising, where $\mathcal{H}(\mathbf{X}):=\mathbf{X}$; 2) inpainting, where $\mathcal{H}(\mathbf{X}):=\mathbf{M}\odot\mathbf{X}$, and $\mathbf{M}$ is a mask matrix indicating missing data; and 3) blurring, where $\mathcal{H}(\mathbf{X}):=\mathbf{K}*\mathbf{X}$, and $\mathbf{K}$ represents convolutional kernels modeling the blur effect. Accordingly, Fig. \ref{fig:analysis}A illustrates representative examples of image restoration results for various degradation scenarios, including denoising, destriping, inpainting, and deblurring.

\subsection*{Image enhancement}
Following restoration, image enhancement further improves the interpretability and analytical utility of HSI data by focusing on spatial and spectral resolution improvements. Spatial enhancement methods \cite{aburaed2023review}, such as pan-sharpening \cite{yokoya2011coupled} and spatial filtering, increase image clarity and facilitate fine-scale feature identification. Spectral enhancement techniques \cite{he2023spectral}, including spectral super-resolution, strengthen spectral discrimination by highlighting subtle yet diagnostically important differences between targets. These complementary strategies collectively maximize visual clarity, diagnostic sensitivity, and analytical reliability. Both enhancement techniques are illustrated in Fig. \ref{fig:analysis}A. Similar to image restoration, the inverse problem of hyperspectral image enhancement can be formulated as follows:
\begin{equation}
\label{eq3}
\begin{aligned}     \mathbf{X}=\mathop{\arg}\mathop{\min}\limits_{\mathbf{X}} \mathcal{L}_{\rm task}(\mathbf{X},\mathbf{Y}) +\lambda \cdot \mathcal{R}_{\rm enhance}(\mathbf{X}),
\end{aligned}
\end{equation} 
where $\mathcal{L}_{\rm task}(\mathbf{X},\mathbf{Y})$ denotes a task-specific loss function quantifying the discrepancy between the target variable $\mathbf{Y}$ and the estimated variable $\mathbf{X}$. The regularization term $\mathcal{R}_{\rm enhance}$ incorporates structural priors relevant to image enhancement, such as edge preservation, frequency domain characteristics, and spatial clarity. 

Hyperspectral enhancement tasks are typically categorized into data fusion and spectral reconstruction. Data fusion methods, such as pansharpening and hyperspectral-multispectral fusion, aim to enhance spatial resolution by integrating complementary high-resolution data sources, e.g., panchromatic (PAN) or multispectral images. A general fusion loss in $\mathcal{L}_{\rm task}$ can be formulated as $\norm{\mathcal{D}(\mathbf{X})-\mathbf{Y}_{\rm LrHS}}^{2}_{\F}+\norm{\mathcal{P}(\mathbf{X})-\mathbf{Y}_{\rm HrPAN}}^{2}_{\F}$, where $\mathcal{D}$ and $\mathcal{P}$ denote spatially and spectrally downsampled operators applied to the estimated fused product $\mathbf{X}$, simulating the acquisition processes of low-resolution HSI data ($\mathbf{Y}_{\rm LrHS}$) and high-resolution PAN images ($\mathbf{Y}_{\rm HrPAN}$), respectively. In contrast, reconstruction methods, such as spectral super-resolution, target the recovery of high-fidelity spectral signatures from degraded sources like multispectral or RGB images. This can be expressed as $\norm{\mathcal{M}(\mathbf{X})-\mathbf{Y}_{\rm HrHS}}^{2}_{\F}$, where $\mathcal{M}$ denotes a learnable spectral upsampling operator, and ($\mathbf{Y}_{\rm HrHS}$) is the reference high-resolution HSI data.

\subsection*{Dimensionality reduction}
\begin{figure*}[!t]
    \centering
    \includegraphics[width=1\linewidth]{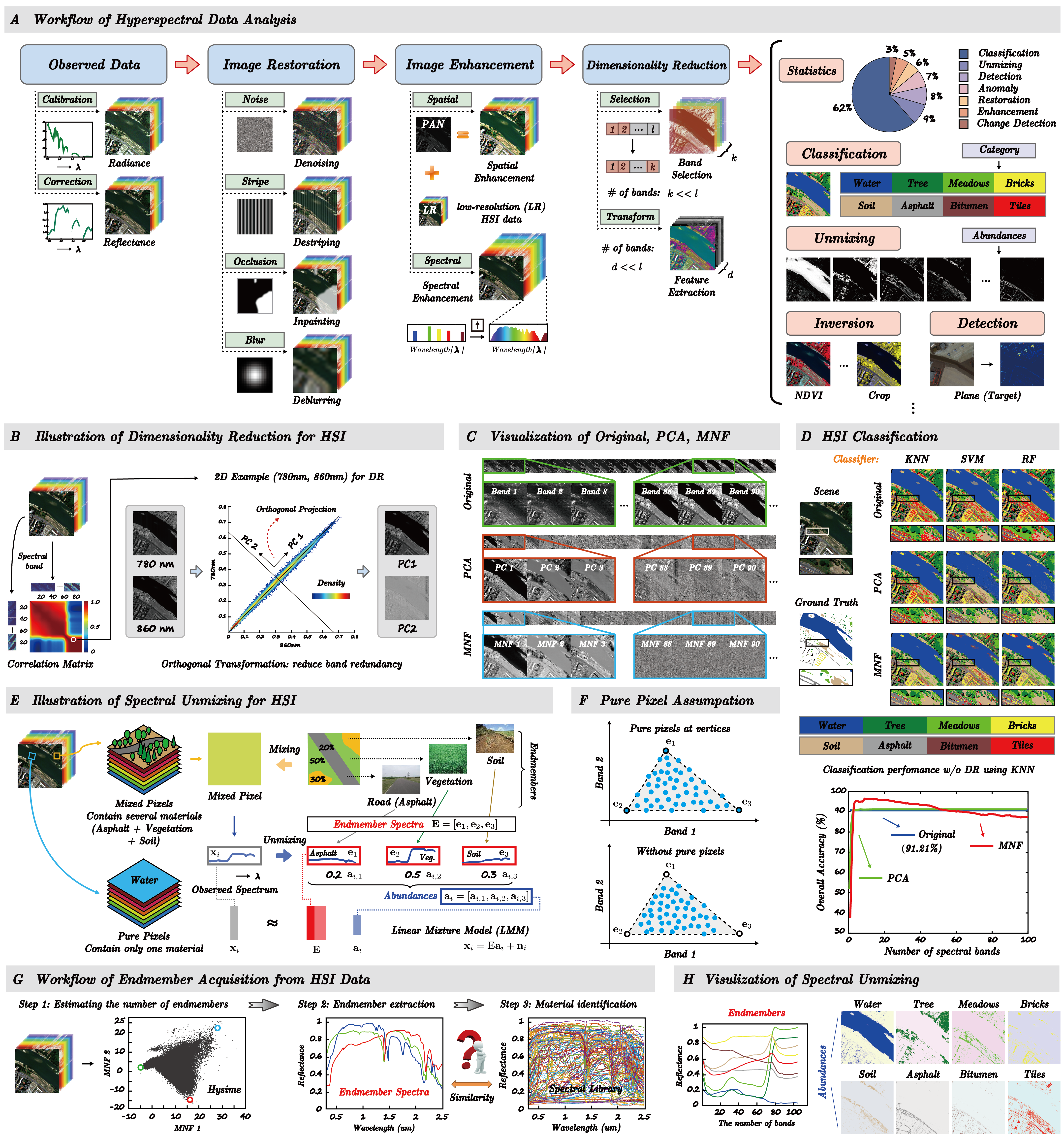}
    \caption{Overview of hyperspectral data analysis, from observed radiance or reflectance HSI data to low-level processing to high-level interpretation. (A) A general workflow of hyperspectral data analysis. (B) Schematic illustration of dimensionality reduction for HSI. (C) Visual comparison of original, PCA-reduced, and MNF-transformed features across three classifiers (KNN, SVM, RF), with magnified views highlighting differences, and a trend plot showing classification accuracy versus the number of spectral bands. (E) Conceptual diagram of spectral unmixing for HSI. (F) Geometric interpretation of the pure pixel assumption, using toy datasets with and without pure pixels. (G) Workflow of endmember acquisition from HSI data. (H) Visualization of unmixing results, including identified or extracted endmembers and corresponding abundance maps for different materials.}
    \label{fig:analysis}
\end{figure*}

After restoration and enhancement, the rich spectral content of HSI enables fine material discrimination owing to hundreds of contiguous bands. However, the inherently high-dimensional characteristics also introduce challenges, including increased computational burden, spectral redundancy, and a heightened risk of overfitting in subsequent analysis tasks. As shown in Fig. \ref{fig:analysis}B, strong inter-band correlation highlights the presence of significant redundancy across spectral dimensions. Dimensionality reduction (DR) is therefore an essential step in the workflow of HSI data analysis \cite{rasti2020feature}, which can be generally categorized into two broad types: band selection and transform-based approaches. As the name suggests, band selection aims to identify a subset of $k$ informative and representative bands (from an original set of $l$ bands) based on certain specific criteria, such as information entropy or inter-band correlation. The band selection process can be formulated using the following discrete expression:
\begin{equation}
\label{eq4}
\begin{aligned}     \mathbf{X}_{\mathcal{S}}=\mathop{\arg}\mathop{\max}\nolimits_{\mathcal{S}\subset \{1,2,...,l\}, \; |\mathcal{S}|=k} \mathcal{J}(\mathbf{X}),
\end{aligned}
\end{equation} 
where $\mathcal{S}$ denotes the subset of selected band indices, with cardinality $|\mathcal{S}|=k$. The resulting data $\mathbf{X}_{\mathcal{S}}$ corresponds to the selected HSI data constructed using the $k$ selected bands from the original image $\mathbf{X}$. The function $\mathcal{J}$ quantifies the informativeness or discriminative capability of the selected band subset. Common choices for $\mathcal{J}$ include mutual information, maximum entropy, and minimum reconstructed error.

Transform-based approaches reduce dimensionality by projecting the original data into a lower-dimensional subspace that preserves the most informative spectral or spatial-spectral structures. Among these, principal component analysis (PCA) \cite{greenacre2022principal} and minimum noise fraction (MNF) \cite{green1988transformation} are two of the most representative methods, often serving as baseline techniques for illustrating DR in HSI due to their easy-interpretable and friendly-practical linearized modeling. More specifically, let $\mathbf{x}_{i}\in \mathbb{R}^{l \times 1}$ be the spectral vector at the $i$-th pixel, where $l$ is the number of spectral bands and $n$ is the total number of pixels in the HSI image. PCA seeks an orthogonal transformation that maximizes the variance of the projected data:
\begin{equation}
\label{eq5}
\begin{aligned}     \mathbf{w}&=\mathop{\arg}\mathop{\max}\limits_{\mathbf{w}} \frac{1}{n}\sum_{i=1}^{n}(\mathbf{w}^{\T}\mathbf{x}_{i}-\mathbf{w}^{\T}\mathbf{\overline{x}})^{2},\\
&=\mathop{\arg}\mathop{\max}\limits_{\mathbf{w}} \mathbf{w}^{\T}\bm{\Sigma}_{\mathbf{x}}\mathbf{w}.
\end{aligned}
\end{equation} 
Here, $\mathbf{w}$ denotes the projection vector, $\mathbf{\overline{x}}$ is the mean spectrum, and $\bm{\Sigma}_{\mathbf{x}}$ is the covariance matrix of the input spectra. The optimization problem in Eq. (\ref{eq5}) boils down to the eigenvalue problem of the variance-covariance matrix, where
\begin{equation}
\label{eq6}
\begin{aligned}     \bm{\Sigma}_{\mathbf{x}}\mathbf{w}=\lambda\mathbf{w}, \;\;\;\; {\rm with} \;\;\;\; \bm{\Sigma}_{\mathbf{x}}=\frac{1}{n}\sum\nolimits^{n}_{i=1}(\mathbf{x}_{i}-\overline{\mathbf{x}})(\mathbf{x}_{i}-\overline{\mathbf{x}})^{\T}.
\end{aligned}
\end{equation} 
This formulation admits a closed-form solution via eigenvalue decomposition, where the leading eigenvectors of $\bm{\Sigma}_{\mathbf{x}}$ define the optimal projection directions for DR. By projecting the original HSI data onto these eigenvectors, PCA captures the most variance-rich components, effectively compressing information while retaining key spectral features. Fig. \ref{fig:analysis}B illustrates the principle of DR for HSI data using a simplified 2D example.

In contrast to PCA, which maximizes total signal variance without explicitly considering noise, MNF transforms the data to maximize the SNR of each component. The MNF transform is effectively a two-stage process: first, noise whitening is performed to decorrelate and normalize noise across bands; second, a PCA-like transformation is applied to the noise-whitened data to identify components ordered by decreasing SNR. Formally, let the observed HSI data be modeled as
\begin{equation}
\label{eq7}
\begin{aligned}
\mathbf{x}_{i}=\mathbf{s}_{i}+\mathbf{n}_{i},
\end{aligned}
\end{equation} 
where $\mathbf{s}_{i}$ is the true underlying signal and $\mathbf{n}_{i}$ is additive noise. Denote $\bm{\Sigma}_{\mathbf{n}}$ and $\bm{\Sigma}_{\mathbf{s}}$ as the covariance matrices of noise and signal, respectively. The MNF transformation solves the generalized eigenvalue problem:
\begin{equation}
\label{eq8}
\begin{aligned}
\bm{\Sigma}_{\mathbf{n}}\mathbf{w}=\lambda\bm{\Sigma}_{\mathbf{s}}\mathbf{w},
\end{aligned}
\end{equation} 
where the eigenvectors ($\mathbf{w}$) define the transform directions and the eigenvalues ($\lambda$) represent the SNRs of the corresponding components. By ordering components in terms of descending SNR, MNF prioritizes information-rich features and suppresses noise-dominated ones, making it particularly well-suited for denoising, anomaly detection, and robust feature extraction in noisy HSI data. Like PCA, MNF also supports an efficient eigendecomposition-based solution, but it explicitly incorporates noise modeling into the projection process.

Representative visualizations before and after DR are presented in Fig. \ref{fig:analysis}C. Comparisons among the original image, PCA-processed output, and MNF-transformed result demonstrate how DR techniques can retain critical information while effectively suppressing noise. PCA tends to concentrate the majority of information into the leading principal components (PCs), although some residual signal may persist in later components. In contrast, MNF achieves improved separation between signal and noise, with most of the meaningful information retained in the early components and minimal residual structure in the latter bands.

\subsection*{Classification}
Classification assigns each pixel to a predefined category or class based on its spectral signature or extracted feature representations \cite{ahmad2025comprehensive}. Classification accounts for approximately 62\% of published HSI-related studies between 2015 and 2025 according to statistics, highlighting its central role in hyperspectral analysis. Broadly, classification methods are categorized as unsupervised or supervised approaches, depending on whether expert-annotated ground-truth labels are available during training. 

Unsupervised classification, often referred to as segmentation, groups pixels into spectrally homogeneous regions without prior class knowledge using clustering algorithms such as K-means \cite{jain1999data}. These methods are useful for exploratory analysis or when ground-truth labels are unavailable, but they typically yield less accurate results in complex scenes. By contrast, supervised classification methods are more prevalent in practical HSI applications and have demonstrated higher accuracy and robustness \cite{rasti2020feature}. These approaches rely on labeled training samples, i.e., spectral inputs paired with class annotations, to train statistical or machine learning models that learn discriminative patterns in the spectral or spatial-spectral domain. Classical supervised algorithms include linear discriminant analysis (LDA) \cite{zhao2024linear}, K-Nearest neighbors (KNN) \cite{peterson2009k}, support vector machines (SVM) \cite{hearst1998support}, and random forests (RF) \cite{breiman2001random}, all of which have been widely applied due to their interpretability and better generalization under well-labeled data regimes. 

Fig. \ref{fig:analysis}D visualizes a typical hyperspectral scene collected over an urban environment scene, its pixel-wise ground-truth annotations, and classification results based on the original HSI data, PCA-reduced features, and MNF-transformed features, using three representative supervised classifiers: KNN, SVM, and RF. In general, these supervised classifiers are trained on a randomly selected subset of labeled pixels and then used to predict class labels of the remaining test samples. Overall accuracy (OA), defined as the proportion of correctly classified pixels relative to the total number of pixels, serves as a standard metric for evaluating classification performance. Overall, DR consistently improves classification performance compared to using full spectral signatures directly. Among the two DR techniques, MNF yields superior accuracy and lower per-class error rates for all pixels, reflecting its strength in noise suppression and informative feature extraction. A locally magnified region further illustrates this trend: classification maps from the original and PCA-reduced data exhibit confusion between similar categories such as Asphalt, Soil, Tree, and Tiles, whereas MNF better preserves class separability more effectively. To further quantify these effects, we evaluate OAs using KNN under varying numbers of spectral bands, with and without DR. PCA achieves performance comparable to the original data, reaching 91.21\% OA with only a few PCs. In contrast, MNF consistently outperforms both, achieving approximately 96\% OA, as noisy or uncertain information is largely suppressed in the leading components, resulting in more discriminative representations.

\subsection*{Spectral unmixing}
HSI systems, more often than not, compromise spatial resolution to yield high spectral resolution with hundreds of narrow (e.g., at 10nm intervals) bands. As a result, spectral mixing commonly occurs at the pixel level, where each pixel typically records a composite signal from multiple underlying materials. Fig. \ref{fig:analysis}E illustrates a macro-level example of this phenomenon. In practice, however, spectral mixing arises not only at macroscopic scales (e.g., adjacent land covers) but also at microscopic levels, where sub-pixel material interactions, such as molecular or even atomic-scale effects, lead to complex mixtures. 

These mixed pixels inevitably degrade the performance of high-level data analysis tasks. Spectral unmixing has therefore emerged as one of the most actively studied topics in HSI, second only to classification (see publication trends in Fig. \ref{fig:analysis}A, around 9\%). Spectral unmixing involves decomposing each observed pixel spectrum into a collection of constituent spectral signatures (i.e., \textit{endmembers}) representing pure materials and their corresponding fractional abundances, typically organized as \textit{abundance maps}. Mathematically, spectral mixing is commonly modeled using a spectral mixture model. As illustrated in Fig. \ref{fig:analysis}E, the linear mixture model (LMM) is widely adopted for simplicity and physical interpretability. LMM assumes a flat surface and single scattering, whereby the observed pixel spectrum is a linear weighted sum of the spectra of pure materials present in the pixel:
\begin{equation}
\label{eq9}
\begin{aligned}
\mathbf{x}_{i}=\mathbf{E}\mathbf{a}_{i}+\mathbf{n}_{i},
\end{aligned}
\end{equation} 
where $\mathbf{E}\in \mathbb{R}^{l\times p}$ is the endmember matrix, with each column representing the spectrum of a pure material, and $\mathbf{a}_{i}\in \mathbb{R}^{p\times 1}$ is the corresponding abundance vector, indicating the proportion of each endmember in pixel $i$. To solve the unmixing problem in Eq. (\ref{eq9}), i.e., to estimate the endmember signatures and their corresponding abundances, two major paradigms are commonly employed: (i) a sequential approach, where endmembers are first extracted (or known) and abundances are then estimated; and (ii) a joint unmixing strategy, which estimates both components simultaneously.

In the sequential approach, endmember spectra are either retrieved from curated spectral libraries or extracted directly from the HSI data. Endmember extraction techniques commonly rely on the pure pixel assumption, which posits that some pixels are predominantly composed of a single material. Fig. \ref{fig:analysis}F provides a geometric interpretation of this assumption, contrasting scenarios with and without pure pixels. In this context, the vertices of the simplex represent pure pixels, or endmembers, each corresponding to a distinct material class. Under this assumption, geometric algorithms such as the Pixel Purity Index (PPI) \cite{boardman1995mapping}, N-FINDR \cite{winter1999n}, and Vertex Component Analysis (VCA) \cite{nascimento2005vertex} are typically used to extract the most spectrally distinct pixels, which are then interpreted as endmembers. The materials corresponding to these endmembers can then be further identified by comparing them with entries in spectral libraries through similarity metrics (e.g., Euclidean distance, spectral angle distance). This overall workflow for endmember extraction and material identification is illustrated in Fig. \ref{fig:analysis}G. Once endmember spectra ($\mathbf{E}$) are defined, the corresponding abundance fractions ($\mathbf{A}=[\mathbf{a}_{1},\mathbf{a}_{2},...,\mathbf{a}_{n}]$) can be estimated by solving a least-squares optimization problem that minimizes the reconstruction error between the observed pixel spectrum ($\mathbf{X}=[\mathbf{x}_{1},\mathbf{x}_{2},...,\mathbf{x}_{n}]$) and its linear approximation:
\begin{equation}
\label{eq10}
\begin{aligned}
\mathbf{a}_{i}&=\mathop{\arg}\mathop{\min}\limits_{\mathbf{a}_{i}}\norm{\mathbf{E}\mathbf{a}_{i}-\mathbf{x}_{i}}^{2}_{2}.
\end{aligned}
\end{equation} 
This unconstrained least-squares problem admits a closed-form analytical solution: $\mathbf{a}_{i}=(\mathbf{E}^{\T}\mathbf{E})^{-1}\mathbf{E}^{\T}\mathbf{x}_{i}$, and serves as the foundation for many abundance estimation techniques. However, to ensure physical plausibility, two constraints are typically imposed on abundances: the \textit{abundance non-negativity constraint (ANC)}, $\mathbf{a}_{i,j}\geq 0$ for all $j$, and the \textit{abundance sum-to-one constraint (ASC)}, $\sum_{j=1}^{p}\mathbf{a}_{i,j}=1$. Under these constraints, the optimization problem becomes:
\begin{equation}
\label{eq11}
\begin{aligned}
\mathbf{a}_{i}&=\mathop{\arg}\mathop{\min}\limits_{\mathbf{a}_{i}}\norm{\mathbf{E}\mathbf{a}_{i}-\mathbf{x}_{i}}^{2}_{2}, \;\; {\rm s.t.} \;\; \mathbf{1}^{\T}\mathbf{a}_{i}=1, \;\; \mathbf{a}_{i}\geq \mathbf{0},
\end{aligned}
\end{equation} 
which can be solved by quadratic programming \cite{nocedal2006quadratic}. Beyond this basic formulation, more advanced models may include additional regularization terms to promote sparsity \cite{bioucas2010alternating} or incorporate prior knowledge from large-scale spectral libraries.

For the joint unmixing strategy, the problem is formulated as a matrix factorization one, also known as blind source separation, which does not rely on the pure pixel assumption and remains applicable even when perfectly pure endmembers are absent from the image. In this setting, both the endmember matrix and the abundance matrix are treated as unknowns to be estimated directly from the observed data. By minimizing the reconstruction error while imposing physical constraints, spectral unmixing reduces to a constrained optimization problem akin to nonnegative matrix factorization (NMF) \cite{lee2000algorithms}:
\begin{equation}
\label{eq12}
\begin{aligned}
\mathbf{A}&=\mathop{\arg}\mathop{\min}\limits_{\mathbf{E},\mathbf{A}}\norm{\mathbf{E}\mathbf{A}-\mathbf{X}}^{2}_{\F}, \;\; {\rm s.t.} \;\; \mathbf{E}\geq \mathbf{0}, \;\; \mathbf{A}\geq \mathbf{0}, \;\; \mathbf{1}^{\T}\mathbf{a}_{i}=1,\;\; i=1,2,...,n.
\end{aligned}
\end{equation} 
This problem is typically solved via alternating optimization, where the endmember matrix $\mathbf{E}$ and the abundance matrix $\mathbf{A}$ are iteratively updated. Initialization of $\mathbf{E}$ is often guided by endmember extraction techniques, such as VCA, to provide physically plausible and spectrally distinct starting points for convergence. Furthermore, the formulation in Eq. (\ref{eq12}) can be flexibly extended by incorporating additional regularization terms (e.g., sparsity, smoothness), allowing priors that reflect domain knowledge or specific application needs. Fig. \ref{fig:analysis}H visualizes estimated results of spectral unmixing, including the extracted or identified endmembers and their corresponding abundance maps for different materials.

\section*{Applications}
\begin{figure*}[!t]
    \centering
    \includegraphics[width=1\linewidth]{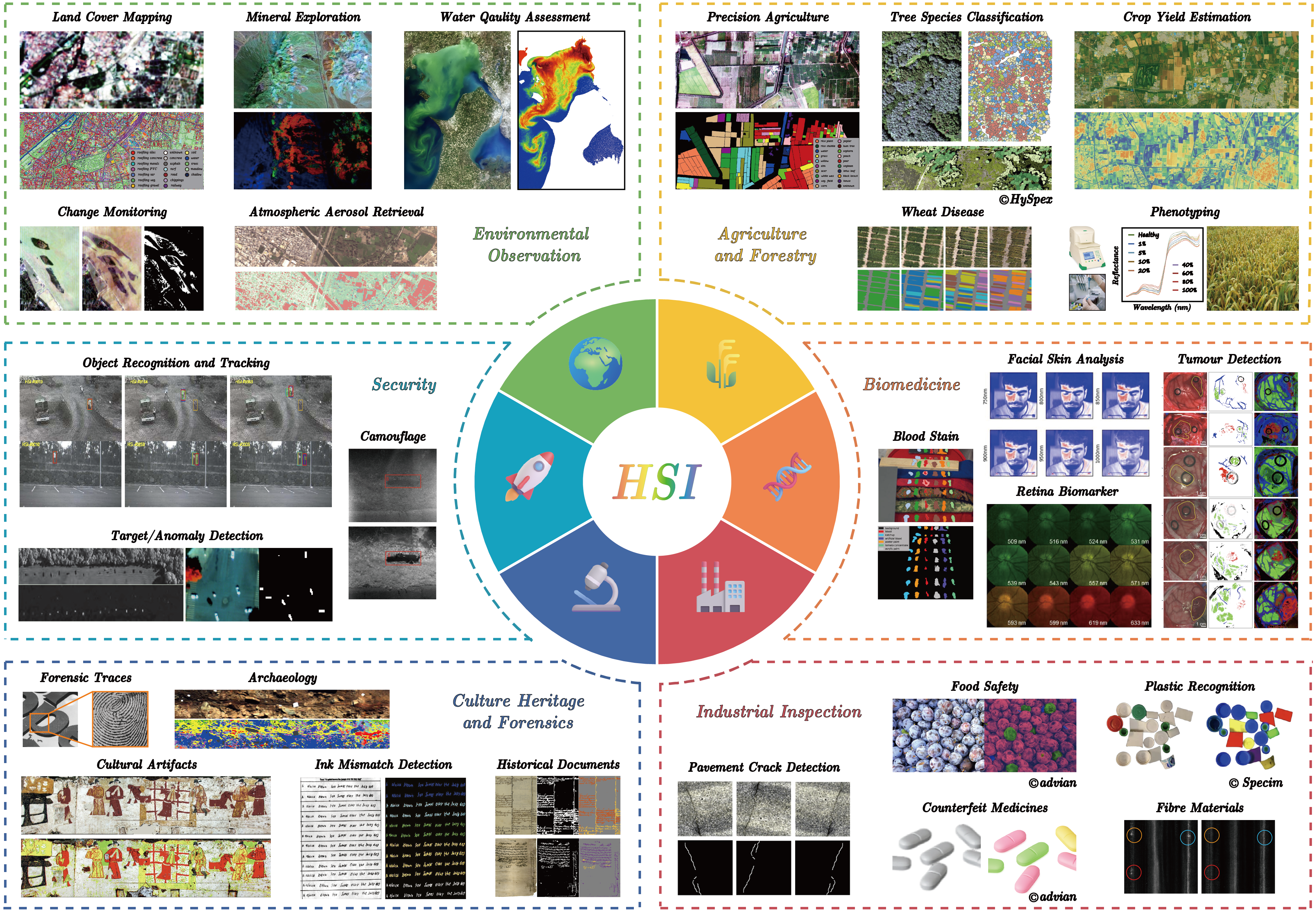}
    \caption{Illustrative examples of HSI applications. Representative use cases of HSI in environmental observation, agriculture and forestry, biomedicine, industrial and food inspection, cultural heritage and forensics, as well as security and defense, are depicted.}
    \label{fig:application}
\end{figure*}

HSI has emerged as a versatile, cross-disciplinary technology that enables fine-grained, non-invasive analysis across diverse domains, from environmental monitoring and agriculture to biomedicine, industrial inspection, cultural heritage, and defense. Several representative examples across different applications are illustrated in Fig. \ref{fig:application}. By capturing continuous spectral signatures at high spatial resolution, HSI allows the identification and quantification of materials, structures, or conditions that are often invisible to conventional imaging modalities. Its ability to operate without physical contact, chemical labeling, or destructive sampling makes it uniquely suited for both \textit{in situ} and remote deployments. Advances in portable and airborne platforms have accelerated the adoption of HSI in dynamic and resource-constrained environments, supporting applications ranging from early disease detection in crops and patients to real-time quality control in manufacturing and autonomous surveillance in complex operational settings. As HSI continues to evolve, its integration with other sensing modalities and its adaptation to embedded, real-time systems are broadening its impact across scientific, industrial, and societal applications.

\subsection*{Environmental observation}
One of the earliest and most impactful domains for HSI has been Earth environmental observation. Leveraging continuous spectral coverage, spaceborne and airborne HSI systems enable precise characterization of land cover, surface dynamics, mineral composition, soil properties, water quality, and atmospheric constituents. These data support fine-grained classification of vegetation types, urban infrastructure, and crop health, facilitating advanced land use analysis \cite{jilge2019gradients}. In geological applications, HSI captures subtle spectral absorption features to map lithological units and identify alteration minerals, aiding resource exploration \cite{van2012multi}. In aquatic environments, it enables the quantification of chlorophyll, turbidity, and dissolved organic matter, supporting early detection of eutrophication and pollution events \cite{Space4Water2024}. Temporal analysis of hyperspectral time series further allows for a detailed monitoring of environmental changes, including deforestation, post-disaster impacts, and urban expansion \cite{ding2025survey}. Moreover, the high spectral fidelity of HSI permits aerosol retrieval and atmospheric profiling \cite{mauceri2019neural}, contributing to improved assessments of air quality and radiative forcing.

\subsection*{Agriculture and forestry}
HSI has emerged as a transformative technology for precision agriculture, plant phenotyping, and forest management, enabling non-invasive, high-resolution monitoring of crop physiology and tree species across spatial and temporal scales. By capturing subtle spectral variations associated with pigment composition, water content, and tissue structure, HSI facilitates fine-grained classification of both crop and tree types \cite{yi2020aerial,ma2024deep}, as well as early detection of biotic and abiotic stressors \cite{thenkabail2019ghisa}, often before visible symptoms appear. These capabilities support timely intervention, optimized input use, and improved system resilience. Time-series HSI further enables dynamic tracking of crop development \cite{yokoya2017multisensor}, allowing key indices—such as chlorophyll content or canopy water status—to be monitored across growth stages and used to model yield potential under varying environmental conditions. When coupled with AI and spectral–spatial modeling, HSI enhances the discrimination of disease severity, nutrient status, and varietal differences in large-scale breeding programs \cite{deng2024rustqnet}. Beyond morphological traits, HSI also detects sub-visual biochemical changes induced by stress or infection, enabling the identification of physiological responses that are imperceptible to the human eye \cite{deng2023quantitative}. This capability supports ultra-early threat detection and informs data-driven breeding strategies through the identification of trait-linked spectral biomarkers.

\subsection*{Biomedicine}
HSI has demonstrated considerable potential in medical diagnostics and surgical guidance by enabling real-time, label-free visualization of tissue composition and pathology. Hyperspectral systems are capable of detecting subtle spectral discrepancies between healthy and abnormal tissues \cite{leon2023hyperspectral}, even when visual cues are absent, due to changes in hemoglobin oxygenation, water content, and cellular morphology. Clinical applications span tumor margin delineation, dermatological lesion classification, blood constituent analysis \cite{ksikazek2020blood}, and retinal biomarker detection.  Intraoperatively, HSI has been explored as a tool to assist surgeons in identifying critical anatomical structures and resection boundaries, reducing reliance on subjective interpretation. In dermatology and ophthalmology, portable HSI systems have been deployed for the classification of skin disorders \cite{ng2023hyper} and the detection of diabetic retinopathy \cite{gao2011snapshot,hadoux2019non}. As a non-ionizing, contrast-free modality, HSI is well suited for repeatable, point-of-care imaging, supporting longitudinal monitoring and early diagnosis in both clinical and outpatient settings.

\subsection*{Industrial inspection}
In manufacturing and food industries, HSI enables rapid, non-destructive assessment of material composition and quality directly on production lines. By capturing spatially resolved spectral signatures, HSI allows precise quantification of key attributes such as sugar and water content, detection of bruising, fungal contamination, or adulterants in food products, and identification of foreign objects in packaging \cite{leiva2013prediction}. These capabilities support both binary classification (e.g., pass/fail) and continuous grading (e.g., ripeness or freshness indices). In pharmaceutical production, HSI is used for verifying active ingredient identity in a non-destructive sampling fashion, assessing coating uniformity, and detecting counterfeit or substandard formulations \cite{coic2019comparison}, thereby enhancing quality assurance. In recycling workflows, HSI sensors are deployed to differentiate and sort plastic waste by polymer type, leveraging subtle spectral differences among materials such as polyethylene terephthalate (PET), high-density polyethylene (HDPE), and polyvinyl chloride (PVC) \cite{zheng2018discrimination}. This enables automated separation with high accuracy, improving recovery efficiency and supporting circular economy goals. Additionally, HSI contributes to flaw detection in composite materials \cite{yan2022non} and infrastructure, such as identifying delamination in fiber-reinforced components or microcracks in pavements \cite{chen2024multiscale}, further extending its utility across industrial quality control domains.

\subsection*{Cultural heritage, forensics, and beyond}
HSI has become an increasingly valuable tool for the non-invasive study of cultural heritage, enabling detailed analysis of cultural artifacts \cite{tang2025digital}, historical documents \cite{lopez2025ink}, and artworks without physical contact or sample preparation. By capturing high-resolution spectral signatures from pigments, binders, and substrates, HSI facilitates the identification of material composition, detection of underdrawings, mapping of retouching, and assessment of degradation or restoration processes. These capabilities are critical for distinguishing authentic works from forgeries, revealing ink mismatches in historical manuscripts \cite{nasir2024hyperspectral}, and informing conservation strategies. In archaeology, HSI supports the remote detection of buried or weathered structures \cite{liang2012advances}, particularly those obscured by vegetation or soil cover. Forensic science also benefits from HSI’s ability to enhance the visibility of latent fingerprints, biological traces, and chemical residues under challenging lighting or background conditions \cite{edelman2012hyperspectral}. Beyond heritage and forensics, emerging applications include smart textiles, autonomous navigation, and recycling systems, where material-specific spectral cues improve classification accuracy, safety, and automation. Collectively, these advances reflect HSI’s transition from a research instrument to a practical, embedded sensing modality across diverse real-world domains.

\subsection*{Security, defense, and surveillance}
HSI has also become an enabling technology in defense and security, owing to its capacity to detect materials and objects based on their unique spectral ``fingerprints''. Unlike conventional RGB or thermal imaging, HSI can uncover subtle spectral anomalies that signal the presence of concealed, disguised, or compositionally distinct targets. This capability is particularly valuable for detecting camouflage \cite{zhao2022camouflage}, synthetic materials, or explosives, and for flagging objects that deviate from typical environmental backgrounds. Anomaly detection algorithms applied to HSI data allow for the discovery of unexpected or suspicious patterns, even in heterogeneous or cluttered scenes where manual inspection fails \cite{nasrabadi2013hyperspectral,li2025learning}. Using airborne or satellite platforms, HSI contributes to wide-area surveillance, such as recognition and tracking \cite{xiong2020material}, providing persistent, data-rich situational awareness in operational settings. Increasingly, HSI is being integrated with complementary sensing modalities \cite{li2024seamo}, such as light detection and ranging (LiDAR), radar, and acoustics, to improve robustness in degraded, contested, or visually complex conditions, reinforcing its strategic role in next-generation multi-sensor intelligence frameworks.

\begin{table*}[!t]
    \centering
    \caption{Representative data repositories for HSI datasets across different domains.}
  \resizebox{1\textwidth}{!}{
    \begin{tabular}{l|c|c|c|c}
        \toprule[1.5pt]
        Dataset & \# of samples & Data variables (size, band, wavelength) & Spectrometer/Sensor & Application \\
        \hline 
        \multirow{2}{*}{Pavia\cite{licciardi2009decision}} & 1 & $610\times 610\times 102$ (430-860nm) & \multirow{2}{*}{ROSIS} & \multirow{2}{*}{Land cover and land use mapping}\\
        & 1 & $1096\times 1096\times 103$ (430-860nm) & & \\
        Cuprite\cite{resmini1997mineral} & 2 & $640\times 320\times 210$ (400-2500nm) & HYDICE & Mineral exploration\\
        Viareggio\cite{acito2016hyperspectral} & 3 & $540\times 375\times 127$ (350-2500nm) & SIM.GA & Change detection\\
        Santa Barbara\cite{lopez2019gpu} & 2 & $984\times 740\times 224$ (400-2500nm) & AVIRIS & Change detection\\
        bayArea Barbara\cite{lopez2019gpu} & 2 & $600\times 500\times 224$ (400-2500nm) & AVIRIS & Change detection\\
        Hermiston city\cite{hasanlou2018hyperspectral} & 2 & $390\times 200\times 242$ (400-2500nm) & HYPERION & Change detection\\
        FWQC\cite{koponen2002lake} & 5,000 & $384\times 384\times 286$ (450-900nm) & AISA & Water quality assessment\\
        AOD-MODIS\cite{wang2011mixture} & 136,400 & $100\times 12$ (400-1440nm) & MODIS & Aerosol Retrieval\\
        AVIRIS-NG\cite{mauceri2019neural} & 21 & $100\times 500\times 319$ (380-2510) & AVIRIS & Aerosol Retrieval\\
        \hline
        Indian Pines\cite{ehu_hyperspectral} & 1 & $145\times 145\times 224$ (400-2500nm) & AVIRIS & Crop classification\\
        Salinas\cite{ehu_hyperspectral} & 1 & $512\times 217\times 224$ (400-2500nm) & AVIRIS & Agricultural land cover\\
        UAV-HSI-Crop\cite{niu2022hsi} & 433 & $96\times 96\times 200$ (385-1021nm) &  Resonon Pika L & Crop type classification\\
        Pest\cite{xiao2022pest} & 52,900 & $9\times 9\times 224$ (400-1000nm) & Specim FX10 & Pest identification\\
        UC-HSI\cite{sankararao2024uc} & 886 & $100\times 100\times 300$ (385-1021nm) & Resonon Pika L & Crop type classification \\
        \multirow{3}{*}{IPK\cite{arend2016quantitative}} & \multirow{3}{*}{10,142} & $2056\times 2454\times 139$ (390-750nm)& Basler Pilot piA2400-17gc & \multirow{3}{*}{Plant Phenotyping}\\
        & & $1234\times 1624\times 152$ (400-750nm)& Basler Scout scA1400-17gc &\\
        & & $254\times 320\times 17$ (1450-1550nm)& Nir 300 PGE & \\
        \hline
        IBT\cite{leon2023hyperspectral} &  61 &  $741\times 1004\times 826$ (400-1000 nm) & Hyperspec & Brain tumour  detection\\
        HyperBlood\cite{romaszewski2021dataset} & 14 & $696\times 520\times 128$ (377-1046 nm) & SOC710 & Blood detection\\
        Hyper-Skin\cite{ng2023hyper} & 330 & $1024\times 1024\times 448$ (400-1000nm)& Specim FX10 & Facial skin analysis\\
        Spectral Retina\cite{falt2011spectral} & 72 & $1024\times 1024\times 30$ (400-700nm) & Canon CR5-45NM & Retinopathy lesion detection\\
        \hline
        GHIFVD\cite{ennis2018hyperspectral} & 42 & $/\times /\times 204$ (376-821nm)& Specim VNIR HS-CL-30-V8E-OEM & Food quality (fruits and vegetables)\\
        \multirow{2}{*}{Chicken Breasts \cite{cai2024effective}} & \multirow{2}{*}{240} & $991\times 960\times 176$(400-1000nm) & Gaiafield-Pro-V10  & \multirow{2}{*}{Food quality (chicken)}\\
        & & $333\times 320\times 256$ (900-1700nm) & GaiaField-Pro-N17E & \\
       Counterfeit Medicines\cite{shinde2020detection} & 24 & $/ \times /\times 213$ (350-1050nm) & MS-720 & Counterfeit medicine detection\\
        Biodegradable Plastics\cite{taneepanichskul2023automatic} & 210 & $/\times /\times 232$ (950-1730nm)& HySpex Baldur S-640i N & Plastic recognition\\
        \multirow{2}{*}{CFRP\cite{yan2022non}} & \multirow{2}{*}{--} & $336\times 336\times 256$ (400-950nm) & Specim V8E & \multirow{2}{*}{Material damage detection}\\
        & & $256\times 320\times 75$ (950-1700nm) & Innospec Red Eye 1.7 & \\
        Expressway\cite{chen2024multiscale} & 1031 & $960\times 1057\times 176$ (394-1001nm) & GaiaSky-mini2 & Pavement crack detection\\
        \hline
        \multirow{2}{*}{Historical Ink\cite{lopez2025ink}} & \multirow{2}{*}{44} & $900\times 900\times 121$ (380-1080nm) & Resonon Pika L & \multirow{2}{*}{Ink classification}\\
        & & $640\times 640\times 161$ (888-1732nm) & Resonon Pika IR+ & \\
        iVision HHID\cite{islam2022ivision} & 270 & $512\times 650\times 149$ (478-901nm) & Imec SNAPSCAN VNIR & Ink mismatch detection\\
        Egyptian Coffin\cite{cucci2024hyperspectral} & 240 & $512\times 512\times 204$ (400-1000nm) & Specim IQ & Cultural artifact recognition\\
        ACSS\cite{gravanis2025descriptor} & 10,000 & $/\times /\times 512$ (400-1050nm) & GER 1500 & Archaeology\\
        BeverageHSI\cite{melit2021forensic} & 1 & $1800\times 11144\times 186$ (400-1000nm) & HySpex VNIR-1800 & Forensic analysis\\
        \hline 
        HOTC\cite{xiong2020material} & 50 & $512\times 256\times 16$ (470-620nm) & XIMEA & Object recognition and tracking\\
        \multirow{2}{*}{ABU\cite{kang2017hyperspectral}} & 12 & $100\times 100\times 188 \sim 150\times 150\times 207$ (400-2500nm) & AVIRIS & \multirow{2}{*}{Target/Anomaly detection}\\
        & 1 & $150\times 150\times 102$ (430-860nm) & ROSIS & \\
        AIR-HAD\cite{li2024interpretable,li2025learning} & 3 & $80\times 100\times 253 \sim 300\times 500\times 253$ (400-1000nm) & SIOM-CAS & Target/Anomaly detection\\
         HOSD\cite{duan2023hyperspectral} & 18 & $869\times 649\times 224 \sim 2302\times 479\times 224$ (365-2500nm) & AVIRIS & Oil spill detection\\
        \bottomrule[1.5pt]
    \end{tabular}
    }
    \label{tab:hsi_datasets}
\end{table*}
        
\section*{Reproducibility and data deposition}
\subsection*{Challenges and community efforts in standardizing HSI practices}
Ensuring reproducibility and enabling responsible data sharing are central to the long-term credibility and usability of HSI in scientific research. However, the complexity and variability inherent to HSI acquisition, preprocessing, and analysis pose significant barriers to consistent results across studies and laboratories. Addressing these issues requires both methodological rigor and community-wide adoption of data standards and repositories.

\subsection*{Issues limiting reproducibility in HSI}
Reproducibility challenges in HSI arise from variability in acquisition setups, sensor configurations, environmental conditions, and inconsistent preprocessing. Minor differences in lighting, viewing angle, or calibration targets can lead to significant changes in spectral response. Replicating results or comparing models becomes infeasible without detailed metadata and transparent processing pipelines. Further issues stem from manual region-of-interest (ROI) selection, underreported annotation criteria, and model tuning on small or homogeneous datasets, which often leads to overfitting. These factors are compounded by the lack of shared protocols for cross-validation, class definitions, or performance metrics.

\subsection*{Community standards and deposition practices}
Efforts to improve reproducibility in HSI are gaining momentum through the establishment of metadata standards and dataset submission guidelines. A well-documented HSI dataset should include: sensor specifications (e.g., spectral range, resolution, band configuration); acquisition metadata (e.g., date, time, location, illumination, weather), calibration details (e.g., white or dark reference, flat-fielding methods), preprocessing steps (e.g., smoothing, normalization, dimensionality reduction), and annotation protocols (e.g., labeling strategy, class definitions, annotator identity). Metadata should be structured in machine-readable formats such as JavaScript Object Notation (JSON), YAML (YAML Ain't Markup Language), or eXtensible Markup Language (XML) to promote interoperability with analysis pipelines. Inclusion of scripts or notebooks for data preprocessing and baseline model training is strongly recommended.

\subsection*{Repositories supporting hyperspectral datasets}
A growing number of repositories support the deposition and dissemination of HSI datasets across diverse scientific domains. Table \ref{tab:hsi_datasets} lists a representative overview of publicly available datasets, annotated by key parameters such as data size, wavelength coverage, sensor type, and applications. These datasets cover EO, agriculture, medicine, cultural heritage, and industrial inspection, reflecting the cross-disciplinary reach of HSI technologies.

To facilitate reproducibility and interoperability, it is recommended that data providers accompany raw and preprocessed files with structured metadata (e.g., acquisition conditions, calibration procedures), label maps or annotations, and clear documentation of processing pipelines. Preferred licenses include open-use frameworks such as Creative Commons Attribution (CC-BY) or Creative Commons Zero (CC0) to ensure broad accessibility and reuse. The adoption of community-agreed data formats and metadata standards will support benchmarking, method comparison, and machine learning model development, enabling robust and generalizable outcomes both in laboratory experiments and field deployments.

\section*{Limitations and optimisations}
HSI has evolved from a niche technology for remote sensing into a versatile tool for scientific, industrial, and societal applications. However, despite substantial progress in sensor design, computational analysis, and integration with AI, HSI remains constrained by several enduring challenges. These limitations stem from the fundamental trade-offs in hardware performance, sensitivity to environmental and acquisition variability, and the complexity of high-dimensional data analysis. Specifically, issues such as the balance between spectral, spatial, and temporal resolution; the reproducibility of measurements across platforms and conditions; and the scalability of algorithms in the face of redundant, noisy, and sparsely annotated data continue to impede broader adoption.

Recent advances are beginning to address these barriers. Innovations in computational hyperspectral imaging, generative preprocessing, pretrained foundation models, and uncertainty quantification are reshaping both acquisition and analysis workflows. These approaches are enabling the development of HSI systems that are more robust, interpretable, and efficient, and that better support deployment in real-world, resource-constrained, and dynamic environments.

\subsection*{Current limitations}
\subsubsection*{Instrumental trade-offs.} HSI systems are fundamentally constrained by several interrelated hardware limitations, stemming from the finite photon budget, sensor throughput, and physical constraints of imaging platforms. These trade-offs shape the design, deployment, and performance of real-world HSI systems:
\begin{itemize}
    \item \textit{Spectral-spatial trade-off}: Increasing spectral resolution typically requires narrower bandwidths and more bands, which reduces the energy available per band. To maintain an adequate SNR, this often comes at the cost of reduced spatial resolution, longer integration times, or increased system complexity.
    \item \textit{SNR vs. acquisition speed}: Achieving high spectral fidelity in real time is challenging, especially in low-light or dynamic environments. Snapshot or fast-scanning systems often suffer from reduced SNR due to shorter dwell times, requiring advanced denoising or signal reconstruction techniques.
    \item \textit{Data volume and transmission bottlenecks}: HSI captures hundreds of contiguous spectral bands, resulting in data volumes several orders of magnitude larger than conventional RGB or multispectral imaging. This places significant demands on memory, bandwidth, and storage, particularly problematic for embedded or UAV-based systems that must transmit or process data in real time.
    \item \textit{Sensor miniaturization vs. performance}: Compact, lightweight HSI systems, necessary for field, drone, or wearable applications, often compromise optical quality, detector sensitivity, or cooling capability. Bridging the gap between benchtop-grade fidelity and mobile operation remains a key engineering challenge.
    \item \textit{Hardware cost and scalability}: High-performance hyperspectral cameras require precision optics, dispersive elements, and sensitive detectors, which remain expensive and difficult to scale. These constraints limit the accessibility of HSI in cost-sensitive domains, such as agriculture or point-of-care diagnostics.
\end{itemize}

\subsubsection*{Acquisition inconsistencies} Real-world HSI is highly susceptible to external physical and environmental conditions, leading to complex spectral variability\cite{borsoi2021spectral,hong2018augmented} that challenges reproducibility, comparability, and cross-domain generalization. These inconsistencies, largely absent in laboratory settings, become critical when transitioning to field-based or spaceborne/airborne applications.
\begin{itemize}
    \item \textit{Illumination variability}: Changes in solar angle, cloud cover, or terrain geometry tend to generate spectral shifts and intensity variations that are difficult to normalize across time or location.
    \item \textit{Atmospheric perturbations}: Water vapor, aerosols, and other atmospheric components introduce nonlinear distortions that differ by wavelength and altitude, particularly affecting SWIR and TIR bands.
    \item \textit{Surface anisotropy}: Material reflectance varies with viewing and illumination angles due to bidirectional reflectance effects, which standard correction methods rarely fully compensate for.
    \item \textit{Sensor-specific characteristics}: Differences in sensor design, calibration procedures, and spectral response functions create nontrivial discrepancies between datasets from different platforms.
    \item \textit{Correction challenges}: Physics-based atmospheric correction and radiometric normalization are difficult to apply robustly across heterogeneous scenes, limiting model transferability and standardization.
\end{itemize}

\subsubsection*{Analytical challenges} The high dimensionality and complex structure of HSI data introduce major hurdles in algorithm design, computational efficiency, and interpretability. These challenges are compounded by variability in data characteristics of sensors and limited availability of high-quality labeled samples.
\begin{itemize}
    \item \textit{High-dimensional structure}: The inherently tensorized spatial-spectral structure of HSI cubes gives rise to high-dimensional and complex data characteristics. When extended to time-series acquisitions, the dimensionality grows even further, intensifying ``the curse of dimensionality'' \cite{koppen2000curse}. This imposes significant challenges for efficient storage, scalable processing, and reliable analysis of the HSI data.
    \item \textit{Spectral redundancy and noise}: The narrow-band acquisition strategy of HSI introduces strong inter-band correlation, resulting in considerable spectral redundancy. When coupled with low SNRs in certain spectral wavelength regions, this significantly hampers the extraction and mining of effective and discriminative information from HSI data.
    \item \textit{Lack of standardization}: Substantial variation in the number of spectral bands, wavelength range, and sensor characteristics across HSI systems hinders the flexible reuse of pretrained models and the reproducibility of benchmark results. This fragmentation limits cross-sensor scalability and poses a major barrier to developing robust, transferable analysis pipelines.
    \item \textit{Limited annotations}: High-quality hyperspectral data analysis relies heavily on accurate ground-truth labeling, yet the annotation process is labor-intensive, time-consuming, and often requires domain expertise. As a result, annotated datasets are typically scarce and limited in scale, which restricts the development of robust models and hinders progress toward large-scale, high-performance, and rapid HSI processing and interpretation.
    \item \textit{Model complexity and opacity}: The high dimensionality and intricate spectral-spatial coupling in HSI data demand specialized model architectures with large computational footprints. While deep learning methods have shown promise, their black-box nature and lack of physical interpretability raise concerns in high-stakes HSI domains such as agriculture, environment, and healthcare, where explainability and trust are essential.
    \item \textit{Domain shifts}: Hyperspectral models trained on laboratory or synthetic datasets often fail to generalize in real-world conditions due to acquisition-specific factors such as lighting variability, atmospheric effects, or material heterogeneity. These domain shifts compromise model robustness and hinder reliable deployment across diverse sensing platforms and application environments.
\end{itemize}

\subsection*{Optimization Strategies}
\subsubsection*{Computational hyperspectral imaging} 
Inherent hardware limitations, such as unavoidable trade-offs between spatial and spectral resolution, data quality, and imaging speed, continue to constrain conventional HSI systems. One promising solution is computational hyperspectral imaging, which tightly integrates optical design with algorithmic reconstruction. By jointly optimizing hardware and digital processing, these hybrid systems aim to overcome traditional performance bottlenecks in HSI.

For example, learned super-resolution or HSI techniques can be embedded directly into the acquisition pipeline \cite{li2024casformer}, enabling simultaneous enhancement of spatial and spectral resolution without compromising SNRs. Similarly, neural architectures\cite{ren2021comprehensive} tailored to hardware constraints, such as lightweight CNNs or quantized spectral encoders, support real-time inference on embedded platforms. Parallel implementation of classification and unmixing algorithms in hardware platforms such as graphics processing units (GPUs) or field programmable gate arrays (FPGAs) is now bridging the gap towards real-time interpretation of hyperspectral data, even onboard the sensing platform \cite{plaza2011parallel}. These synergistic designs bridge the gap between compact, low-cost sensors and the performance requirements of high-fidelity applications, representing a key direction for the next generation of HSI systems.

\subsubsection*{Generative preprocessing} Hyperspectral data acquisition is inevitably affected by a combination of physical, chemical, environmental, and instrumental factors, which challenge the reproducibility, reliability, and transferability of high-level analyses. Traditional physics-based correction models, although valuable, often fall short of comprehensively addressing these challenges, particularly in the presence of nonlinear, scene-dependent distortions and sensor-specific artifacts. In contrast, generative AI models \cite{feuerriegel2024generative}, such as adversarial learning, style transfer, and sensor-aware normalization, can learn to model and compensate for such complexity and variability for data recovery and reconstruction in a data-driven fashion. These models have demonstrated enhanced adaptability and generalization for tasks such as data recovery and spectral reconstruction. Integrating generative models into the HSI processing pipeline as a form of generative preprocessing represents a powerful and promising strategy for mitigating acquisition inconsistencies and improving the robustness of subsequent data analyses.

\subsubsection*{Pretrained foundation models} Given the high-dimensional and complex structure of HSI data, the challenge of extracting discriminant information from redundant and noisy spectra, and the scarcity of labeled samples, pretrained foundation models are emerging as a powerful solution. Transformer-based architectures \cite{dosovitskiy2020image}, for example, can be trained on large-scale HSI datasets in a self-supervised manner \cite{liu2021self}, enabling the extraction of rich spectral-spatial representations from unlabeled inputs. These models, once pretrained, require only minimal labeled data for fine-tuning, yet can achieve performance levels comparable to models trained on substantially larger annotated datasets. This ``pretraining + fine-tuning'' paradigm is becoming the mainstream strategy in HSI\cite{li2024s2mae,hong2024spectralgpt}, offering a flexible and scalable approach to support multiple downstream tasks with improved generalization and data efficiency.

\subsubsection*{Uncertainty quantification} Although data-driven machine learning approaches are increasingly applied to HSI, concerns remain regarding their stability, transferability, and practical reliability. This is largely because such models typically learn representations entirely from data in a black-box manner, often leading to opacity and unpredictable behavior, particularly when deployed in new scenes or domains. These uncertainties can be quantified or reduced through explicit confidence modeling, domain knowledge integration, or Bayesian frameworks\cite{gawlikowski2023survey}, offering a promising path to mitigate these challenges. Such strategies can enhance trustworthiness, reproducibility, and transparency\cite{li2024innovation}, which are critical for sensitive or high-stakes HSI applications in fields such as agriculture, healthcare, and environmental monitoring.

\section*{Outlook}
Building on the current limitations and emerging optimisation strategies in HSI, we outline a long-term vision for the future of the field. This outlook highlights key directions that could shape the next generation of HSI, integrating advances in hardware, computational models, and community-driven initiatives.

\subsection*{3J vision: future hyperspectral imaging, processing, and analysis}
Future advances in HSI will likely be shaped by integrative strategies that unify optical design, data processing, and analytical modeling holistically and synergistically. We propose the 3J vision as a conceptual framework for next-generation HSI systems: joint design of software and hardware, joint utilization of diverse HSI sensors, and joint modeling of HSI-centered multimodal data. This idea emphasizes the tight coupling of sensing, computation, quantization, and interpretation, aiming to overcome current HSI limitations in spatial-spectral trade-offs, data standardization, and cross-domain generalization. By embracing this vision, future HSI platforms could achieve unprecedented performance, scalability, and adaptability across scientific, industrial, and societal applications.

\subsubsection*{The first J: joint design of HSI software and hardware} 
Traditional HSI development often treats sensor hardware and data analysis pipelines as independent components, leading to suboptimal end-to-end performance and mismatches between acquisition constraints and algorithmic needs. Joint design promotes the co-optimization of hardware (e.g., optics, detectors, scanning mechanisms) and software (e.g., reconstruction algorithms, AI models, quantum learning) to achieve superior overall system efficiency.

\subsubsection*{The second J: joint utilization of different HSI sensors} 
Future HSI systems will increasingly benefit from the coordinated use of multiple hyperspectral sensors across different platforms (e.g., satellite, airborne, UAV, handheld) and wavelength ranges (e.g., VNIR, SWIR, TIR). By jointly leveraging their complementary spatial, spectral, and temporal characteristics, such systems can achieve multi-scale, multi-resolution sensing that overcomes the limitations of any single sensor type. This joint utilization supports more complete scene understanding, robust cross-validation, and improved generalization across diverse environments. Realizing this vision requires advances in sensor fusion algorithms, harmonization of spectral responses, and unified calibration protocols to ensure consistent data quality across platforms.

\subsubsection*{The third J: joint modeling of HSI-centered multiple modalities} 
The next generation of HSI analytics will increasingly rely on joint modeling frameworks that integrate hyperspectral data with complementary sensing modalities, such as depth sensors (e.g., LiDAR or structured light), radar systems (e.g., SAR), thermal infrared cameras, event cameras, or even semantic and contextual data (e.g., text, metadata, or social media data). This multimodal fusion allows HSI to serve as a central layer in a richer, multi-source perception system, providing enhanced depth, structural context, and material specificity that can not be obtained from spectral information alone. Such joint models hold promise for overcoming current limitations in spatial resolution, depth estimation, and semantic interpretability. Key challenges include the development of cross-modal alignment algorithms, unified feature representations, and scalable architectures that can efficiently process and learn from these heterogeneous data sources cohesively.

\subsection*{Structuralized HSI: towards omni-view imaging}
Although HSI data is represented as a 3D cube (two spatial dimensions and one spectral dimension), it is essentially a pseudo-3D structure. Current HSI techniques typically rely on a single or fixed imaging viewpoint, yielding a flat projection of the scene without providing true stereoscopic or volumetric information. Future developments could push HSI beyond this limitation toward multi-view and structuralized imaging, integrating principles of light field imaging with hyperspectral data capture to reconstruct true 3D or even 4D spectral-spatial representations (adding temporal dynamics and obtaining hyperspectral videos).

By combining multi-angle or multi-aperture imaging systems with computational reconstruction algorithms, one could obtain both spectral and geometric depth information simultaneously, leading to richer scene understanding. Generative AI models will play a crucial role in this vision by learning to synthesize volumetric hyperspectral cubes from sparse viewpoints, interpolate missing angles, or even reconstruct unseen perspectives. This approach would result in omni-view HSI, a holistic representation of objects and environments with complete spectral and geometric fidelity.

Such structuralized HSI has transformative potential in applications like digital twins, precision robotics, cultural heritage preservation, and autonomous navigation, where both material composition and 3D geometry are critical. In the longer term, fusing light field imaging, structured illumination, and hyperspectral sensing with AI-driven modeling could create a new generation of 4D HSI platforms that unify spectral, spatial, angular, and temporal dimensions into a single, comprehensive sensing framework.

\subsection*{HSI Foundation models and agents: a vision for ``HSI brain''}

Driven by the rapid progress of AI, HSI is evolving toward foundation models that act as a universal ``Spectral Brain'', capable of handling diverse tasks, sensors, and environments within a single framework. These models are trained on large-scale, heterogeneous datasets, learning holistic spectral-spatial representations that unify imaging, preprocessing, analysis, and interpretation. By embedding physics-informed priors and self-supervised spectral learning, HSI foundation models aim to overcome persistent bottlenecks such as data sparsity, sensor-specific fragmentation, and domain shifts.

Looking ahead, these foundation models can serve as the core intelligence for HSI agents, i.e., autonomous systems that can perceive their environment, adaptively analyze spectral cues, and make context-aware decisions in real time. An HSI agent, powered by such a model, would seamlessly integrate sensing, reasoning, and action: from dynamically adjusting acquisition parameters, to performing onboard analysis, to generating actionable insights for applications ranging from precision agriculture to autonomous planetary exploration. This vision positions the HSI agent as a self-optimizing, universally deployable platform, redefining how hyperspectral data are captured and utilized.

\subsection*{Future HSI: one for all}
The explosive development of AI and its spin-off technologies is poised to fundamentally reshape what is possible in HSI. Looking ahead over the next 5 to 10 years, we envision a transformative one-for-all (OFA) paradigm for HSI, which is a conceptual framework that redefines both the data structure and analytical model of hyperspectral systems. 

On the data side, \textbf{``OFA data''} envisions super-extremely compressive HSI representations, in which a full hyperspectral cube could be reduced to a compact, information-rich, single-band equivalent. Such a representation would enable the reconstruction of any desired spectral cube, feature map, or analytical product on demand. This approach could break longstanding limitations related to sensor type, data size, resolution, or platform specificity, offering unprecedented advantages for data storage, transmission, and cross-sensor interoperability. By alleviating hardware burdens, OFA data could enable scalable, adaptive, and universally accessible HSI analytics, suitable for both resource-rich and resource-constrained environments.

On the model side, the OFA outlook can also extend to the modeling domain through the concept of an \textbf{``OFA models''}. As the name suggests, an OFA model aims to unify disparate HSI processing tasks and inputs within a single, flexible architecture. The OFA model would integrate diverse data types, spanning different spectral ranges, spatial resolutions, or acquisition modalities, also consolidate multiple analytical models or pipelines into one framework. In effect, regardless of the number or nature of inputs, a single OFA model could adaptively process and analyze them, eliminating the need for task-specific or sensor-specific models. This concept points to a future in which hyperspectral analytics are streamlined, universally applicable, and inherently scalable across domains and applications.

On the application side, we foresee the emergence of \textbf{``OFA applications''}, in which a single intelligent system can flexibly address diverse hyperspectral use cases, from environmental monitoring and precision agriculture to healthcare, industry, and cultural heritage, without requiring customized models for each scenario. This ``OFA'' approach would streamline workflows, accelerate deployment, and democratize access to hyperspectral analytics, ultimately shaping a universal platform for science, technology, and real-world decision-making.

\subsection*{Community efforts for open, responsible, and scalable HSI}
The future impact of HSI will depend not only on technological advances but also on sustained community-driven efforts to promote openness, reproducibility, and accessibility. Building shared infrastructure, including open-source toolkits, benchmark datasets, annotation platforms, and model repositories, is essential to foster transparency, comparability, and cross-domain progress. Equally important is lowering technical barriers to adoption through modular software libraries, cloud-based analysis platforms, and domain-specific training resources. These efforts will empower researchers, practitioners, and decision-makers in fields as diverse as agriculture, ecology, environmental monitoring, and healthcare. As HSI technologies become increasingly integrated into operational systems, community leadership will play a critical role in ensuring responsible development, rigorous validation, and equitable access to spectral intelligence across global contexts.

\bibliography{sample}

@ARTICLE{goetz1985imaging,
  title={Imaging spectrometry for earth remote sensing},
  author={Goetz, Alexander FH and Vane, Gregg and Solomon, Jerry E and Rock, Barrett N},
  journal={Science},
  volume={228},
  number={4704},
  pages={1147--1153},
  year={1985},
  publisher={American Association for the Advancement of Science}
}

@article{vane1993airborne,
  title={The airborne visible/infrared imaging spectrometer (AVIRIS)},
  author={Vane, Gregg and Green, Robert O and Chrien, Thomas G and Enmark, Harry T and Hansen, Earl G and Porter, Wallace M},
  journal={Remote Sensing of Environment},
  volume={44},
  number={2-3},
  pages={127--143},
  year={1993},
  publisher={Elsevier}
}

@misc{chlus2025hytools,
  author = {Chlus, Adam and Ye, Zhe and Zheng, Tian and Queally, Nicole and Greenberg, Eric and Townsend, Philip},
  title = {{HyTools (1.6.0)}},
  year = {2025},
  publisher = {Zenodo},
  url = {https://doi.org/10.5281/zenodo.15240407},
  note = {Accessed: 2025-04-28}
}

@misc{boggs2020spectral,
  author = {Boggs, Thomas},
  title = {{Spectral Python (SPy) User Guide}},
  year = {2020},
  howpublished = {\url{https://www.spectralpython.net/user_guide.html}},
  note = {Accessed: 2023-11-24}
}

@article{bakker2024hyperspectral,
  title={Hyperspectral Python: HypPy},
  author={Bakker, Wim and van Ruitenbeek, Frank and van der Werff, Harald and Hecker, Christoph and Dijkstra, Arjan and van der Meer, Freek},
  journal={Algorithms},
  volume={17},
  number={8},
  pages={337},
  year={2024},
  publisher={MDPI}
}

@article{greenacre2022principal,
  title={Principal component analysis},
  author={Greenacre, Michael and Groenen, Patrick JF and Hastie, Trevor and d’Enza, Alfonso Iodice and Markos, Angelos and Tuzhilina, Elena},
  journal={Nature Reviews Methods Primers},
  volume={2},
  number={1},
  pages={100},
  year={2022},
  publisher={Nature Publishing Group UK London}
}

@article{green1988transformation,
  title={A transformation for ordering multispectral data in terms of image quality with implications for noise removal},
  author={Green, Andrew A and Berman, Mark and Switzer, Paul and Craig, Maurice D},
  journal={IEEE Transactions on Geoscience and Remote Sensing},
  volume={26},
  number={1},
  pages={65--74},
  year={1988},
  publisher={IEEE}
}

@article{jain1999data,
  title={Data clustering: a review},
  author={Jain, Anil K and Murty, M Narasimha and Flynn, Patrick J},
  journal={ACM Computing Surveys},
  volume={31},
  number={3},
  pages={264--323},
  year={1999},
  publisher={Acm New York, NY, USA}
}

@article{zhao2024linear,
  title={Linear discriminant analysis},
  author={Zhao, Shuping and Zhang, Bob and Yang, Jian and Zhou, Jianhang and Xu, Yong},
  journal={Nature Reviews Methods Primers},
  volume={4},
  number={1},
  pages={70},
  year={2024},
  publisher={Nature Publishing Group UK London}
}

@article{hearst1998support,
  title={Support vector machines},
  author={Hearst, Marti A. and Dumais, Susan T and Osuna, Edgar and Platt, John and Scholkopf, Bernhard},
  journal={IEEE Intelligent Systems and Their Applications},
  volume={13},
  number={4},
  pages={18--28},
  year={1998},
  publisher={IEEE}
}

@article{breiman2001random,
  title={Random forests},
  author={Breiman, Leo},
  journal={Machine Learning},
  volume={45},
  pages={5--32},
  year={2001},
  publisher={Springer}
}

@article{peterson2009k,
  title={K-nearest neighbor},
  author={Peterson, Leif E},
  journal={Scholarpedia},
  volume={4},
  number={2},
  pages={1883},
  year={2009}
}

@article{rasti2020feature,
  title={Feature extraction for hyperspectral imagery: The evolution from shallow to deep: Overview and toolbox},
  author={Rasti, Behnood and Hong, Danfeng and Hang, Renlong and Ghamisi, Pedram and Kang, Xudong and Chanussot, Jocelyn and Benediktsson, Jon Atli},
  journal={IEEE Geoscience and Remote Sensing Magazine},
  volume={8},
  number={4},
  pages={60--88},
  year={2020},
  publisher={IEEE}
}

@inproceedings{winter1999n,
  title={N-FINDR: An algorithm for fast autonomous spectral end-member determination in hyperspectral data},
  author={Winter, Michael E},
  booktitle={Imaging spectrometry V},
  volume={3753},
  pages={266--275},
  year={1999},
  organization={SPIE}
}

@inproceedings{boardman1995mapping,
  title={Mapping target signatures via partial unmixing of AVIRIS data},
  author={Boardman, Joseph W and Kruse, Fred A and Green, Robert O},
  booktitle={The fifth annual JPL Airborne Earth Science Workshop.},
  year={1995},
  pages={23--26},
  address = {Pasadena, CA}
}

@article{nocedal2006quadratic,
  title={Quadratic programming},
  author={Nocedal, Jorge and Wright, Stephen J},
  journal={Numerical optimization},
  pages={448--492},
  year={2006},
  publisher={Springer}
}

@inproceedings{bioucas2010alternating,
  title={Alternating direction algorithms for constrained sparse regression: Application to hyperspectral unmixing},
  author={Bioucas-Dias, Jos{\'e} M and Figueiredo, M{\'a}rio AT},
  booktitle={2010 2nd Workshop on Hyperspectral Image and Signal Processing: Evolution in Remote Sensing},
  pages={1--4},
  year={2010},
  organization={IEEE}
}

@article{lee2000algorithms,
  title={Algorithms for non-negative matrix factorization},
  author={Lee, Daniel and Seung, H Sebastian},
  journal={Advances in neural information processing systems},
  volume={13},
  year={2000}
}

@article{jilge2019gradients,
  title={Gradients in urban material composition: A new concept to map cities with spaceborne imaging spectroscopy data},
  author={Jilge, Marianne and Heiden, Uta and Neumann, Carsten and Feilhauer, Hannes},
  journal={Remote Sensing of Environment},
  volume={223},
  pages={179--193},
  year={2019},
  publisher={Elsevier}
}

@article{hong2018augmented,
  title={An augmented linear mixing model to address spectral variability for hyperspectral unmixing},
  author={Hong, Danfeng and Yokoya, Naoto and Chanussot, Jocelyn and Zhu, Xiao Xiang},
  journal={IEEE Transactions on Image Processing},
  volume={28},
  number={4},
  pages={1923--1938},
  year={2018},
  publisher={IEEE}
}

@misc{Space4Water2024,
  author       = {{Space4Water}},
  title        = {Exploring the Exciting Potential of Hyperspectral Imaging for Water Quality Monitoring},
  year         = {2022},
  howpublished = {\url{https://www.space4water.org/news/exploring-exciting-potential-hyperspectral-imaging-water-quality-monitoring}},
  note={Retrieved 2025-06-01}
}

@article{ding2025survey,
  title={A Survey of Sample-Efficient Deep Learning for Change Detection in Remote Sensing: Tasks, strategies, and challenges},
  author={Ding, Lei and Hong, Danfeng and Zhao, Maofan and Chen, Hongruixuan and Li, Chenyu and Deng, Jie and Yokoya, Naoto and Bruzzone, Lorenzo and Chanussot, Jocelyn},
  journal={IEEE Geoscience and Remote Sensing Magazine},
  year={2025},
  publisher={IEEE}
}

@article{mauceri2019neural,
  title={Neural network for aerosol retrieval from hyperspectral imagery},
  author={Mauceri, Steffen and Kindel, Bruce and Massie, Steven and Pilewskie, Peter},
  journal={Atmospheric Measurement Techniques},
  volume={12},
  number={11},
  pages={6017--6036},
  year={2019}
}

@article{deng2024rustqnet,
  title={RustQNet: Multimodal deep learning for quantitative inversion of wheat stripe rust disease index},
  author={Deng, Jie and Hong, Danfeng and Li, Chenyu and Yao, Jing and Yang, Ziqian and Zhang, Zhijian and Chanussot, Jocelyn},
  journal={Computers and Electronics in Agriculture},
  volume={225},
  pages={109245},
  year={2024},
  publisher={Elsevier}
}

@article{yi2020aerial,
  title={Aerial hyperspectral remote sensing classification dataset of Xiongan New Area (Matiwan Village)},
  author={Yi, CEN and Zhang, Lifu and Zhang, Xia and Yueming, WANG and Wenchao, QI and Senlin, TANG and Zhang, Peng},
  journal={National Remote Sensing Bulletin},
  volume={24},
  number={11},
  pages={1299--1306},
  year={2020},
  publisher={National Remote Sensing Bulletin}
}

@online{thenkabail2019ghisa,
  author       = {Thenkabail, Prasad and Aneece, I.},
  title        = {Global Hyperspectral Imaging Spectral-library of Agricultural crops for Conterminous United States V001},
  year         = {2019},
  url          = {https://doi.org/10.5067/Community/GHISA/GHISACONUS.001},
  note         = {Distributed by NASA EOSDIS Land Processes Distributed Active Archive Center},
  urldate      = {2025-06-01}
}

@article{yokoya2017multisensor,
  title={Multisensor coupled spectral unmixing for time-series analysis},
  author={Yokoya, Naoto and Zhu, Xiao Xiang and Plaza, Antonio},
  journal={IEEE Transactions on Geoscience and Remote Sensing},
  volume={55},
  number={5},
  pages={2842--2857},
  year={2017},
  publisher={IEEE}
}

@article{deng2023quantitative,
  title={Quantitative estimation of wheat stripe rust disease index using unmanned aerial vehicle hyperspectral imagery and innovative vegetation indices},
  author={Deng, Jie and Wang, Rui and Yang, Lujia and Lv, Xuan and Yang, Ziqian and Zhang, Kai and Zhou, Congying and Pengju, Li and Wang, Zhifang and Abdullah, Ahsan and others},
  journal={IEEE Transactions on Geoscience and Remote Sensing},
  volume={61},
  pages={1--11},
  year={2023},
  publisher={IEEE}
}

@article{leon2023hyperspectral,
  title={Hyperspectral imaging benchmark based on machine learning for intraoperative brain tumour detection},
  author={Leon, Raquel and Fabelo, Himar and Ortega, Samuel and Cruz-Guerrero, Ines A and Campos-Delgado, Daniel Ulises and Szolna, Adam and Pi{\~n}eiro, Juan F and Espino, Carlos and O’Shanahan, Aruma J and Hernandez, Maria and others},
  journal={npj Precision Oncology},
  volume={7},
  number={1},
  pages={119},
  year={2023},
  publisher={Nature Publishing Group UK London}
}

@article{ng2023hyper,
  title={Hyper-skin: A hyperspectral dataset for reconstructing facial skin-spectra from RGB images},
  author={Ng, Pai Chet and Chi, Zhixiang and Verdie, Yannick and Lu, Juwei and Plataniotis, Konstantinos N},
  journal={Advances in Neural Information Processing Systems},
  volume={36},
  pages={24158--24170},
  year={2023}
}

@article{ksikazek2020blood,
  title={Blood stain classification with hyperspectral imaging and deep neural networks},
  author={Ksiazek, Kamil and Romaszewski, Michal and Glomb, Przemyslaw and Grabowski, Bartosz and Cholewa, Michal},
  journal={Sensors},
  volume={20},
  number={22},
  pages={6666},
  year={2020},
  publisher={MDPI}
}

@article{hadoux2019non,
  title={Non-invasive in vivo hyperspectral imaging of the retina for potential biomarker use in Alzheimer’s disease},
  author={Hadoux, Xavier and Hui, Flora and Lim, Jeremiah KH and Masters, Colin L and P{\'e}bay, Alice and Chevalier, Sophie and Ha, Jason and Loi, Samantha and Fowler, Christopher J and Rowe, Christopher and others},
  journal={Nature Communications},
  volume={10},
  number={1},
  pages={4227},
  year={2019},
  publisher={Nature Publishing Group UK London}
}

@article{gao2011snapshot,
  title={Snapshot hyperspectral retinal camera with the Image Mapping Spectrometer (IMS)},
  author={Gao, Liang and Smith, R Theodore and Tkaczyk, Tomasz S},
  journal={Biomedical optics express},
  volume={3},
  number={1},
  pages={48--54},
  year={2011},
  publisher={Optical Society of America}
}

@article{yan2022non,
  title={Non-destructive testing of composite fiber materials with hyperspectral imaging—Evaluative studies in the EU H2020 FibreEUse project},
  author={Yan, Yijun and Ren, Jinchang and Zhao, Huan and Windmill, James FC and Ijomah, Winifred and De Wit, Jesper and Von Freeden, Justus},
  journal={IEEE Transactions on Instrumentation and Measurement},
  volume={71},
  pages={1--13},
  year={2022},
  publisher={IEEE}
}

@article{leiva2013prediction,
  title={Prediction of firmness and soluble solids content of blueberries using hyperspectral reflectance imaging},
  author={Leiva-Valenzuela, Gabriel A and Lu, Renfu and Aguilera, Jos{\'e} Miguel},
  journal={Journal of Food Engineering},
  volume={115},
  number={1},
  pages={91--98},
  year={2013},
  publisher={Elsevier}
}

@article{chen2024multiscale,
  title={A multiscale enhanced pavement crack segmentation network coupling spectral and spatial information of UAV hyperspectral imagery},
  author={Chen, Xiao and Zhang, Xianfeng and Ren, Miao and Zhou, Bo and Sun, Min and Feng, Ziyuan and Chen, Baoying and Zhi, Xiaobo},
  journal={International Journal of Applied Earth Observation and Geoinformation},
  volume={128},
  pages={103772},
  year={2024},
  publisher={Elsevier}
}

@article{zheng2018discrimination,
  title={A discrimination model in waste plastics sorting using NIR hyperspectral imaging system},
  author={Zheng, Yan and Bai, Jiarui and Xu, Jingna and Li, Xiayang and Zhang, Yimin},
  journal={Waste Management},
  volume={72},
  pages={87--98},
  year={2018},
  publisher={Elsevier}
}

@article{coic2019comparison,
  title={Comparison of hyperspectral imaging techniques for the elucidation of falsified medicines composition},
  author={Coic, Laureen and Sacr{\'e}, Pierre-Yves and Dispas, Amandine and Sakira, Abdoul Karim and Fillet, Marianne and Marini, Roland D and Hubert, Philippe and Ziemons, Eric},
  journal={Talanta},
  volume={198},
  pages={457--463},
  year={2019},
  publisher={Elsevier}
}

@article{liang2012advances,
  title={Advances in multispectral and hyperspectral imaging for archaeology and art conservation},
  author={Liang, Haida},
  journal={Applied Physics A},
  volume={106},
  pages={309--323},
  year={2012},
  publisher={Springer}
}

@article{nasir2024hyperspectral,
  title={A hyperspectral unmixing approach for ink mismatch detection in unbalanced clusters},
  author={Nasir, Faryal Aurooj and Liaquat, Salman and Khurshid, Khurram and Mahyuddin, Nor Muzlifah},
  journal={Journal of Information and Intelligence},
  volume={2},
  number={2},
  pages={177--190},
  year={2024},
  publisher={Elsevier}
}

@article{edelman2012hyperspectral,
  title={Hyperspectral imaging for non-contact analysis of forensic traces},
  author={Edelman, Gerda J and Gaston, Edurne and Van Leeuwen, Ton G and Cullen, PJ and Aalders, Maurice CG},
  journal={Forensic science international},
  volume={223},
  number={1-3},
  pages={28--39},
  year={2012},
  publisher={Elsevier}
}

@article{tang2025digital,
  title={Digital restoration of mural paintings from late Tang tomb M1373 in Xi’an based on hyperspectral analysis and image interaction processing},
  author={Tang, Xingjia and Yan, Jing and Zhang, Pengchang and Dong, Wenqiang and He, Zhang and Qiu, Shi and Zeng, Zimu},
  journal={npj Heritage Science},
  volume={13},
  number={1},
  pages={1--22},
  year={2025},
  publisher={Nature Publishing Group}
}

@article{lopez2025ink,
  title={Ink classification in historical documents using hyperspectral imaging and machine learning methods},
  author={L{\'o}pez-Baldomero, Ana Bel{\'e}n and Buzzelli, Marco and Moronta-Montero, Francisco and Mart{\'\i}nez-Domingo, Miguel {\'A}ngel and Valero, Eva Mar{\'\i}a},
  journal={Spectrochimica Acta Part A: Molecular and Biomolecular Spectroscopy},
  volume={335},
  pages={125916},
  year={2025},
  publisher={Elsevier}
}

@inproceedings{zhao2022camouflage,
  title={Camouflage target recognition based on dimension reduction analysis of hyperspectral image regions},
  author={Zhao, Jiale and Zhou, Bing and Wang, Guanglong and Liu, Jie and Ying, Jiaju},
  booktitle={Photonics},
  volume={9},
  number={9},
  pages={640},
  year={2022},
  organization={MDPI}
}

@article{nasrabadi2013hyperspectral,
  title={Hyperspectral target detection: An overview of current and future challenges},
  author={Nasrabadi, Nasser M},
  journal={IEEE Signal Processing Magazine},
  volume={31},
  number={1},
  pages={34--44},
  year={2013},
  publisher={IEEE}
}

@article{xiong2020material,
  title={Material based object tracking in hyperspectral videos},
  author={Xiong, Fengchao and Zhou, Jun and Qian, Yuntao},
  journal={IEEE Transactions on Image Processing},
  volume={29},
  pages={3719--3733},
  year={2020},
  publisher={IEEE}
}

@article{li2024seamo,
  title={SeaMo: A Season-Aware Multimodal Foundation Model for Remote Sensing},
  author={Li, Xuyang and Li, Chenyu and Vivone, Gemine and Hong, Danfeng},
  journal={Information Fusion},
  year={2024}
}

@article{koponen2002lake,
  title={Lake water quality classification with airborne hyperspectral spectrometer and simulated MERIS data},
  author={Koponen, Sampsa and Pulliainen, Jouni and Kallio, Kari and Hallikainen, Martti},
  journal={Remote Sensing of Environment},
  volume={79},
  number={1},
  pages={51--59},
  year={2002},
  publisher={Elsevier}
}

@article{lopez2019gpu,
  title={GPU framework for change detection in multitemporal hyperspectral images},
  author={L{\'o}pez-Fandi{\~n}o, Javier and B. Heras, Dora and Arg{\"u}ello, Francisco and Dalla Mura, Mauro},
  journal={International Journal of Parallel Programming},
  volume={47},
  number={2},
  pages={272--292},
  year={2019},
  publisher={Springer}
}

@article{acito2016hyperspectral,
  title={Hyperspectral airborne “Viareggio 2013 Trial” data collection for detection algorithm assessment},
  author={Acito, Nicola and Matteoli, Stefania and Rossi, Alessandro and Diani, Marco and Corsini, Giovanni},
  journal={IEEE Journal of Selected Topics in Applied Earth Observations and Remote Sensing},
  volume={9},
  number={6},
  pages={2365--2376},
  year={2016},
  publisher={IEEE}
}

@article{xiao2022pest,
  title={Pest identification via hyperspectral image and deep learning},
  author={Xiao, Zhitao and Yin, Kai and Geng, Lei and Wu, Jun and Zhang, Fang and Liu, Yanbei},
  journal={Signal, Image and Video Processing},
  volume={16},
  number={4},
  pages={873--880},
  year={2022},
  publisher={Springer}
}

@article{arend2016quantitative,
  title={Quantitative monitoring of Arabidopsis thaliana growth and development using high-throughput plant phenotyping},
  author={Arend, Daniel and Lange, Matthias and Pape, Jean-Michel and Weigelt-Fischer, Kathleen and Arana-Ceballos, Fernando and M{\"u}cke, Ingo and Klukas, Christian and Altmann, Thomas and Scholz, Uwe and Junker, Astrid},
  journal={Scientific Data},
  volume={3},
  number={1},
  pages={1--13},
  year={2016},
  publisher={Nature Publishing Group}
}

@article{sankararao2024uc,
  title={UC-HSI: UAV Based Crop Hyperspectral Imaging Datasets and Machine Learning Benchmark Results},
  author={Sankararao, Adduru UG and Rajalakshmi, P and Choudhary, Sunita},
  journal={IEEE Geoscience and Remote Sensing Letters},
  year={2024},
  publisher={IEEE}
}

@article{romaszewski2021dataset,
  title={A dataset for evaluating blood detection in hyperspectral images},
  author={Romaszewski, Micha{\l} and G{\l}omb, Przemys{\l}aw and Sochan, Arkadiusz and Cholewa, Micha{\l}},
  journal={Forensic science international},
  volume={320},
  pages={110701},
  year={2021},
  publisher={Elsevier}
}

@article{falt2011spectral,
  title={Spectral imaging of the human retina and computationally determined optimal illuminants for diabetic retinopathy lesion detection},
  author={F{\"a}lt, Pauli and Hiltunen, Jouni and Hauta-Kasari, Markku and Sorri, Iiris and Kalesnykiene, Valentina and Pietil{\"a}, Juhani and Uusitalo, Hannu},
  journal={J. Imaging Sci. Technol},
  volume={55},
  number={3},
  pages={030509},
  year={2011}
}

@article{ennis2018hyperspectral,
  title={Hyperspectral database of fruits and vegetables},
  author={Ennis, Robert and Schiller, Florian and Toscani, Matteo and Gegenfurtner, Karl R},
  journal={Journal of the Optical Society of America A},
  volume={35},
  number={4},
  pages={B256--B266},
  year={2018},
  publisher={OSA}
}

@article{cai2024effective,
  title={An effective deep learning fusion method for predicting the TVB-N and TVC contents of chicken breasts using dual hyperspectral imaging systems},
  author={Cai, Mingrui and Li, Xiaoxin and Liang, Juntao and Liao, Ming and Han, Yuxing},
  journal={Food Chemistry},
  volume={456},
  pages={139847},
  year={2024},
  publisher={Elsevier}
}

@article{taneepanichskul2023automatic,
  title={Automatic identification and classification of compostable and biodegradable plastics using hyperspectral imaging},
  author={Taneepanichskul, Nutcha and Hailes, Helen C and Miodownik, Mark},
  journal={Frontiers in Sustainability},
  volume={4},
  pages={1125954},
  year={2023},
  publisher={Frontiers Media SA}
}

@inproceedings{shinde2020detection,
  title={Detection of counterfeit medicines using hyperspectral sensing},
  author={Shinde, Sujit R and Bhavsar, Karan and Kimbahune, Sanjay and Khandelwal, Sundeep and Ghose, Avik and Pal, Arpan},
  booktitle={2020 42nd Annual International Conference of the IEEE Engineering in Medicine \& Biology Society (EMBC)},
  pages={6155--6158},
  year={2020},
  organization={IEEE}
}

@article{islam2022ivision,
  title={iVision HHID: Handwritten hyperspectral images dataset for benchmarking hyperspectral imaging-based document forensic analysis},
  author={Islam, Ammad Ul and Khan, Muhammad Jaleed and Asad, Muhammad and Khan, Haris Ahmad and Khurshid, Khurram},
  journal={Data in Brief},
  volume={41},
  pages={107964},
  year={2022},
  publisher={Elsevier}
}

@article{cucci2024hyperspectral,
  title={Hyperspectral imaging and convolutional neural networks for augmented documentation of ancient Egyptian artefacts},
  author={Cucci, Costanza and Guidi, Tommaso and Picollo, Marcello and Stefani, Lorenzo and Python, Lorenzo and Argenti, Fabrizio and Barucci, Andrea},
  journal={Heritage Science},
  volume={12},
  number={1},
  pages={75},
  year={2024},
  publisher={Springer}
}

@article{melit2021forensic,
  title={Forensic analysis of beverage stains using hyperspectral imaging},
  author={Melit Devassy, Binu and George, Sony},
  journal={Scientific Reports},
  volume={11},
  number={1},
  pages={6512},
  year={2021},
  publisher={Nature Publishing Group UK London}
}

@article{kang2017hyperspectral,
  title={Hyperspectral anomaly detection with attribute and edge-preserving filters},
  author={Kang, Xudong and Zhang, Xiangping and Li, Shutao and Li, Kenli and Li, Jun and Benediktsson, J{\'o}n Atli},
  journal={IEEE Transactions on Geoscience and Remote Sensing},
  volume={55},
  number={10},
  pages={5600--5611},
  year={2017},
  publisher={IEEE}
}

@article{duan2023hyperspectral,
  title={Hyperspectral remote sensing benchmark database for oil spill detection with an isolation forest-guided unsupervised detector},
  author={Duan, Puhong and Kang, Xudong and Ghamisi, Pedram and Li, Shutao},
  journal={IEEE Transactions on Geoscience and Remote Sensing},
  volume={61},
  pages={1--11},
  year={2023},
  publisher={IEEE}
}

@article{li2024interpretable,
  title={Interpretable networks for hyperspectral anomaly detection: A deep unfolding solution},
  author={Li, Chenyu and Zhang, Bing and Hong, Danfeng and Yao, Jing and Jia, Xiuping and Plaza, Antonio and Chanussot, Jocelyn},
  journal={IEEE Transactions on Geoscience and Remote Sensing},
  year={2024},
  publisher={IEEE}
}

@article{li2025learning,
  title={Learning disentangled priors for hyperspectral anomaly detection: A coupling model-driven and data-driven paradigm},
  author={Li, Chenyu and Zhang, Bing and Hong, Danfeng and Jia, Xiuping and Plaza, Antonio and Chanussot, Jocelyn},
  journal={IEEE Transactions on Neural Networks and Learning Systems},
  year={2025},
  publisher={IEEE}
}

@article{niu2022hsi,
  title={HSI-TransUNet: A transformer based semantic segmentation model for crop mapping from UAV hyperspectral imagery},
  author={Niu, Bowen and Feng, Quanlong and Chen, Boan and Ou, Cong and Liu, Yiming and Yang, Jianyu},
  journal={Computers and Electronics in Agriculture},
  volume={201},
  pages={107297},
  year={2022},
  publisher={Elsevier}
}

@article{bioucas2012hyperspectral,
  title={Hyperspectral unmixing overview: Geometrical, statistical, and sparse regression-based approaches},
  author={Bioucas-Dias, Jos{\'e} M and Plaza, Antonio and Dobigeon, Nicolas and Parente, Mario and Du, Qian and Gader, Paul and Chanussot, Jocelyn},
  journal={IEEE journal of selected topics in applied earth observations and remote sensing},
  volume={5},
  number={2},
  pages={354--379},
  year={2012},
  publisher={IEEE}
}

@article{ahmad2025comprehensive,
  title={A comprehensive survey for hyperspectral image classification: The evolution from conventional to transformers and mamba models},
  author={Ahmad, Muhammad and Distefano, Salvatore and Khan, Adil Mehmood and Mazzara, Manuel and Li, Chenyu and Li, Hao and Aryal, Jagannath and Ding, Yao and Vivone, Gemine and Hong, Danfeng},
  journal={Neurocomputing},
  pages={130428},
  year={2025},
  publisher={Elsevier}
}

@article{hong2021interpretable,
  title={Interpretable hyperspectral artificial intelligence: When nonconvex modeling meets hyperspectral remote sensing},
  author={Hong, Danfeng and He, Wei and Yokoya, Naoto and Yao, Jing and Gao, Lianru and Zhang, Liangpei and Chanussot, Jocelyn and Zhu, Xiaoxiang},
  journal={IEEE Geoscience and Remote Sensing Magazine},
  volume={9},
  number={2},
  pages={52--87},
  year={2021},
  publisher={IEEE}
}

@article{hong2024spectralgpt,
  title={SpectralGPT: Spectral remote sensing foundation model},
  author={Hong, Danfeng and Zhang, Bing and Li, Xuyang and Li, Yuxuan and Li, Chenyu and Yao, Jing and Yokoya, Naoto and Li, Hao and Ghamisi, Pedram and Jia, Xiuping and others},
  journal={IEEE Transactions on Pattern Analysis and Machine Intelligence},
  volume={46},
  number={8},
  pages={5227--5244},
  year={2024},
  publisher={IEEE}
}

@inproceedings{li2024s2mae,
  title={S2mae: A spatial-spectral pretraining foundation model for spectral remote sensing data},
  author={Li, Xuyang and Hong, Danfeng and Chanussot, Jocelyn},
  booktitle={Proceedings of the IEEE/CVF Conference on Computer Vision and Pattern Recognition},
  pages={24088--24097},
  year={2024}
}

@book{rybicki2024radiative,
  title={Radiative processes in astrophysics},
  author={Rybicki, George B and Lightman, Alan P},
  year={2024},
  publisher={John Wiley \& Sons}
}

@article{hardeberg2002multispectral,
  title={Multispectral color image capture using a liquid crystal tunable filter},
  author={Hardeberg, Jon Y and Schmitt, Francis and Brettel, Hans},
  journal={Optical engineering},
  volume={41},
  number={10},
  pages={2532--2548},
  year={2002},
  publisher={Society of Photo-Optical Instrumentation Engineers}
}

@article{chang1981acousto,
  title={Acousto-optic tunable filters},
  author={Chang, IC},
  journal={Optical Engineering},
  volume={20},
  number={6},
  pages={824--829},
  year={1981},
  publisher={SPIE}
}

@article{wang2011mixture,
  title={Mixture model for multiple instance regression and applications in remote sensing},
  author={Wang, Zhuang and Lan, Liang and Vucetic, Slobodan},
  journal={IEEE Transactions on Geoscience and Remote Sensing},
  volume={50},
  number={6},
  pages={2226--2237},
  year={2011},
  publisher={IEEE}
}

@article{licciardi2009decision,
  title={Decision fusion for the classification of hyperspectral data: Outcome of the 2008 GRS-S data fusion contest},
  author={Licciardi, Giorgio and Pacifici, Fabio and Tuia, Devis and Prasad, Saurabh and West, Terrance and Giacco, Ferdinando and Thiel, Christian and Inglada, Jordi and Christophe, Emmanuel and Chanussot, Jocelyn and others},
  journal={IEEE Transactions on Geoscience and Remote Sensing},
  volume={47},
  number={11},
  pages={3857--3865},
  year={2009},
  publisher={IEEE}
}

@article{resmini1997mineral,
  title={Mineral mapping with hyperspectral digital imagery collection experiment (HYDICE) sensor data at Cuprite, Nevada, USA},
  author={Resmini, RG and Kappus, ME and Aldrich, WS and Harsanyi, JC and Anderson, M},
  journal={International Journal of Remote Sensing},
  volume={18},
  number={7},
  pages={1553--1570},
  year={1997},
  publisher={Taylor \& Francis}
}

@misc{ehu_hyperspectral,
  author       = {{Universidad del País Vasco}},
  title        = {Hyperspectral Remote Sensing Scenes},
  year         = {2022},
  url          = {https://www.ehu.eus/ccwintco/index.php/Hyperspectral_Remote_Sensing_Scenes},
  note         = {Accessed: 2025-06-12}
}

@article{dosovitskiy2020image,
  title={An image is worth 16x16 words: Transformers for image recognition at scale},
  author={Dosovitskiy, Alexey and Beyer, Lucas and Kolesnikov, Alexander and Weissenborn, Dirk and Zhai, Xiaohua and Unterthiner, Thomas and Dehghani, Mostafa and Minderer, Matthias and Heigold, Georg and Gelly, Sylvain and others},
  journal={arXiv preprint arXiv:2010.11929},
  year={2020}
}

@article{liu2021self,
  title={Self-supervised learning: Generative or contrastive},
  author={Liu, Xiao and Zhang, Fanjin and Hou, Zhenyu and Mian, Li and Wang, Zhaoyu and Zhang, Jing and Tang, Jie},
  journal={IEEE Transactions on Knowledge and Data Engineering},
  volume={35},
  number={1},
  pages={857--876},
  year={2021},
  publisher={IEEE}
}

@article{li2024casformer,
  title={CasFormer: Cascaded transformers for fusion-aware computational hyperspectral imaging},
  author={Li, Chenyu and Zhang, Bing and Hong, Danfeng and Zhou, Jun and Vivone, Gemine and Li, Shutao and Chanussot, Jocelyn},
  journal={Information Fusion},
  volume={108},
  pages={102408},
  year={2024},
  publisher={Elsevier}
}

@article{ren2021comprehensive,
  title={A comprehensive survey of neural architecture search: Challenges and solutions},
  author={Ren, Pengzhen and Xiao, Yun and Chang, Xiaojun and Huang, Po-Yao and Li, Zhihui and Chen, Xiaojiang and Wang, Xin},
  journal={ACM Computing Surveys (CSUR)},
  volume={54},
  number={4},
  pages={1--34},
  year={2021},
  publisher={ACM New York, NY, USA}
}

@inproceedings{koppen2000curse,
  title={The curse of dimensionality},
  author={K{\"o}ppen, Mario},
  booktitle={5th online world conference on soft computing in industrial applications (WSC5)},
  volume={1},
  pages={4--8},
  year={2000}
}

@article{feuerriegel2024generative,
  title={Generative ai},
  author={Feuerriegel, Stefan and Hartmann, Jochen and Janiesch, Christian and Zschech, Patrick},
  journal={Business \& Information Systems Engineering},
  volume={66},
  number={1},
  pages={111--126},
  year={2024},
  publisher={Springer}
}

@article{gawlikowski2023survey,
  title={A survey of uncertainty in deep neural networks},
  author={Gawlikowski, Jakob and Tassi, Cedrique Rovile Njieutcheu and Ali, Mohsin and Lee, Jongseok and Humt, Matthias and Feng, Jianxiang and Kruspe, Anna and Triebel, Rudolph and Jung, Peter and Roscher, Ribana and others},
  journal={Artificial Intelligence Review},
  volume={56},
  number={Suppl 1},
  pages={1513--1589},
  year={2023},
  publisher={Springer}
}

@article{li2024innovation,
  title={Interpretable foundation models as decryptors peering into the Earth system},
  author={Li, Chenyu and Hong, Danfeng and Zhang, Bing and Liao, Tianjun and Yokoya, Naoto and Ghamisi, Pedram and Chen, Min and Wang, Lizhe and Benediktsson, Jon Atli and Chanussot, Jocelyn},
  journal={The Innovation},
  volume={5},
  number={5},
  year={2024},
  publisher={Elsevier}
}

@article{borsoi2021spectral,
  title={Spectral variability in hyperspectral data unmixing: A comprehensive review},
  author={Borsoi, Ricardo Augusto and Imbiriba, Tales and Bermudez, Jos{\'e} Carlos Moreira and Richard, C{\'e}dric and Chanussot, Jocelyn and Drumetz, Lucas and Tourneret, Jean-Yves and Zare, Alina and Jutten, Christian},
  journal={IEEE Geoscience and Remote Sensing Magazine},
  volume={9},
  number={4},
  pages={223--270},
  year={2021},
  publisher={IEEE}
}

@article{hong2024multimodal,
  title={Multimodal artificial intelligence foundation models: Unleashing the power of remote sensing big data in Earth observation},
  author={Hong, Danfeng and Li, Chenyu and Zhang, Bing and Yokoya, Naoto and Benediktsson, Jon Atli and Chanussot, Jocelyn},
  journal={The Innovation Geoscience},
  volume={2},
  number={1},
  pages={100055},
  year={2024},
  publisher={The Innovation Geoscience}
}

@article{van2012multi,
  title={Multi-and hyperspectral geologic remote sensing: A review},
  author={Van der Meer, Freek D and Van der Werff, Harald MA and Van Ruitenbeek, Frank JA and Hecker, Chris A and Bakker, Wim H and Noomen, Marleen F and Van Der Meijde, Mark and Carranza, E John M and De Smeth, J Boudewijn and Woldai, Tsehaie},
  journal={International journal of applied Earth observation and geoinformation},
  volume={14},
  number={1},
  pages={112--128},
  year={2012},
  publisher={Elsevier}
}

@article{yokoya2011coupled,
  title={Coupled nonnegative matrix factorization unmixing for hyperspectral and multispectral data fusion},
  author={Yokoya, Naoto and Yairi, Takehisa and Iwasaki, Akira},
  journal={IEEE Transactions on Geoscience and Remote Sensing},
  volume={50},
  number={2},
  pages={528--537},
  year={2011},
  publisher={IEEE}
}

@article{aburaed2023review,
  title={A review of spatial enhancement of hyperspectral remote sensing imaging techniques},
  author={Aburaed, Nour and Alkhatib, Mohammed Q and Marshall, Stephen and Zabalza, Jaime and Al Ahmad, Hussain},
  journal={IEEE Journal of Selected Topics in Applied Earth Observations and Remote Sensing},
  volume={16},
  pages={2275--2300},
  year={2023},
  publisher={IEEE}
}

@article{he2023spectral,
  title={Spectral super-resolution meets deep learning: Achievements and challenges},
  author={He, Jiang and Yuan, Qiangqiang and Li, Jie and Xiao, Yi and Liu, Denghong and Shen, Huanfeng and Zhang, Liangpei},
  journal={Information Fusion},
  volume={97},
  pages={101812},
  year={2023},
  publisher={Elsevier}
}

@article{nascimento2005vertex,
  title={Vertex component analysis: A fast algorithm to unmix hyperspectral data},
  author={Nascimento, Jos{\'e} MP and Dias, Jos{\'e} MB},
  journal={IEEE Transactions on Geoscience and Remote Sensing},
  volume={43},
  number={4},
  pages={898--910},
  year={2005},
  publisher={IEEE}
}

@article{hasanlou2018hyperspectral,
  title={Hyperspectral change detection: An experimental comparative study},
  author={Hasanlou, Mahdi and Seydi, Seyd Teymoor},
  journal={International journal of remote sensing},
  volume={39},
  number={20},
  pages={7029--7083},
  year={2018},
  publisher={Taylor \& Francis}
}

@article{plaza2011parallel,
  title={Parallel hyperspectral image and signal processing [applications corner]},
  author={Plaza, Antonio and Plaza, Javier and Paz, Abel and Sanchez, Sergio},
  journal={IEEE Signal Processing Magazine},
  volume={28},
  number={3},
  pages={119--126},
  year={2011},
  publisher={IEEE}
}

@article{gravanis2025descriptor,
  title={Descriptor: Archaeological cropmark synthetic signatures (ACSS)},
  author={Gravanis, Elias and Agapiou, Athos},
  journal={IEEE Data Descriptions},
  year={2025},
  publisher={IEEE}
}

@book{chang2003hyperspectral,
  title={Hyperspectral imaging: techniques for spectral detection and classification},
  author={Chang, Chein-I},
  volume={1},
  year={2003},
  publisher={Springer Science \& Business Media}
}

@article{ma2024deep,
  title={A deep-learning-based tree species classification for natural secondary forests using unmanned aerial vehicle hyperspectral images and LiDAR},
  author={Ma, Ye and Zhao, Yuting and Im, Jungho and Zhao, Yinghui and Zhen, Zhen},
  journal={Ecological Indicators},
  volume={159},
  pages={111608},
  year={2024},
  publisher={Elsevier}
}

\end{document}